\documentclass[letterpaper, 10 pt, journal, twoside]{IEEEtran} % Comment this line out if you need a4paper

\IEEEoverridecommandlockouts                              % This command is only needed if 
\usepackage{moreverb,url}

\usepackage[colorlinks,bookmarksopen,bookmarksnumbered,citecolor=red,urlcolor=red]{hyperref}

\usepackage{graphicx} % for \includegraphics
\DeclareGraphicsExtensions{.pdf} % for omit extensio

\usepackage{amsmath} % assumes amsmath package installed
\usepackage[short]{optidef} % for argmini environment
\usepackage[ruled,linesnumbered]{algorithm2e} % for pseudocode

\usepackage{caption} % for figure caption skip setting

\usepackage{bm} % for \bm

\usepackage{multirow}

\usepackage{times} % assumes new font selection scheme installed
\usepackage{color,soul} % for highlighting the word
\usepackage{color} % for \color
\usepackage{tikz}
\usepackage{pgfplots}
\usetikzlibrary{spy,calc}
\usepackage{amssymb} % assumes amsmath package installed

\usepackage{amsmath}

\usepackage{xcolor}
\usepackage{lineno}
\usepackage[normalem]{ulem}
\renewcommand\hl{} % comment this line to turn on highlights
\newcommand\BibTeX{{\rmfamily B\kern-.05em \textsc{i\kern-.025em b}\kern-.08em
		T\kern-.1667em\lower.7ex\hbox{E}\kern-.125emX}}

\newtheorem{remark}{Remark}
\newcommand{\mytilde}{\raise.17ex\hbox{$\scriptstyle\mathtt{‌​\sim}$}}

\usepackage[flushleft]{threeparttable}
\usepackage{tabularx,booktabs}
\usepackage{color} % for \color
\usepackage[mathscr]{eucal}
\pgfplotsset{compat=1.18}
\usepackage{amssymb}  % assumes amsmath package installed
\newcommand{\stacked}[1]{{#1}^{\bm{\mathscr{S}}}}

\newcommand{\JJ}{\mathscr{J}}

\newcommand{\seqJ}[2][]{%
	\IfEqCase{#1}{
		{g}{\stacked{\JJ}_{#2}}    %general
		{h}{\stacked{\widehat{\JJ}}_{#2}}    %hat
	}[\PackageError{currT}{Undefined option to tree: #1}{}]
}

\newcommand{\JS}[2]{%
	\IfEqCase{#1}{
		{g}{\stacked{\JJ}}    %general
		{h}{\stacked{\widehat{\JJ}}}    %hat
	}[\PackageError{currT}{Undefined option to tree: #1}{}]
}

\newcommand{\timeIndex}[1]{
	\IfEqCase{#1}{
		{g}{i}%    %general
		{d}{i}%    %dynamics
		{m}{j}%    %measurement
	}[\PackageError{timeIndex}{Undefined option to tree: #1}{}]
}  
\newcommand{\currT}[1]{%
	\IfEqCase{#1}{
		{g}{\timeIndex{g}}    %general
		{d}{\timeIndex{d}}    %dynamics
		{m}{\timeIndex{m}}    %measurement
	}[\PackageError{currT}{Undefined option to tree: #1}{}]
}
\newcommand{\nextT}[1]{
	\IfEqCase{#1}{%
		{g}{\timeIndex{g}+1}    %general
		{d}{\timeIndex{d}+1}    %dynamics
		{m}{\timeIndex{m}+1}    %measurement
	}[\PackageError{\nextT}{Undefined option to tree: #1}{}]
}%
\newcommand{\prevT}[1]{
	\IfEqCase{#1}{%
		{g}{\timeIndex{g}-1}    %general
		{d}{\timeIndex{d}-1}    %dynamics
		{m}{\timeIndex{m}-1}    %measurement
	}[\PackageError{\prevT}{Undefined option to tree: #1}{}]
}%

\usepackage{xstring}
\title{
   Superellipse Function: A Powerful Geometric Tool for the Design and Deformation Modeling of Tendon-Driven Continuum Manipulators With Flexural Patterns Experiencing Tension Loss }
\author{Yang Liu$^{1}$ and Farshid Alambeigi$^{2}$,\IEEEmembership{~Member,~IEEE}
\thanks{*Research reported in this publication was supported by the National Institute of Biomedical Imaging and Bioengineering of the National Institutes of Health under Award Number R21EB030796.}
	\thanks{$^{1}$Y. Liu is with the Walker Department of Mechanical Engineering at the University of Texas at Austin, Austin, TX, 78712, USA. Email: {\tt\footnotesize liu.yang@utexas.edu}.}
    \thanks{$^{2}$F.~Alambeigi is with the Department of Mechanical Engineering at Rice University, and The Bookout Center at Houston Methodist Research Institute. Email: {\tt\footnotesize alambeigi@rice.edu}.}
}
\usepackage{color} % for \textcolor

\usepackage[linesnumbered, ruled]{algorithm2e}
\SetKwRepeat{Do}{do}{while}
\let\oldnl\nl% Store \nl in \oldnl
\newcommand{\nonl}{\renewcommand{\nl}{\let\nl\oldnl}}% Remove line number for one line

\begin{document}
	
	\maketitle
	%%%%%%%%%%%%%%%%%%%%%%%%%%%%%%%%%%%%%%%%%%%%%%%%%%%%%%%%%%%%%%%%%%%%%%%%%%%%%%%%
	\begin{abstract}
	In this article, we feature the \textit{Superellipse function}
as a  geometric and versatile tool
for effective design and modeling of flexural patterns (FPs) in  tendon-driven continuum manipulators (TD-CMs) with  discontinuous cross-section geometries. We also propose a unique complementary mathematical framework and solution algorithm for  modeling the effects of distributed and concentrated  friction forces--  caused by partially-constrained tendons-- on tension loss and deformation behavior of a TD-CM designed using a Superellipse function. 
Particularly, by employing this geometric function, we establish four intuitive functional  mappings relating the manipulator's geometric parameters, kinematics, and deformation behavior.
Thanks to the features of these intuitive mappings and Superellipse function, we then introduce the concepts of \textit{design space} and \textit{design curve} to facilitate intuitive design of a  TD-CM with a generic FP geometry  experiencing tendon tension loss. 
To thoroughly evaluate the performance of this novel framework and its aforementioned features, we performed various simulation and experimental studies on TD-CMs designed and fabricated based on a Superellipse function. Results were then analyzed and compared  using different error metrics. 
	\end{abstract}
	%%%%%%%%%%%%%%%%%%%%%%%%%%%%%%%%%%%%%%%%%%%%%%%%%%%%%%%%%%%%%%%%%%%%%%%%%%%%%%%%
	\section{Introduction} \label{section:intro}
\begin{figure}[!t]
\centering
\includegraphics[width=0.9\columnwidth]{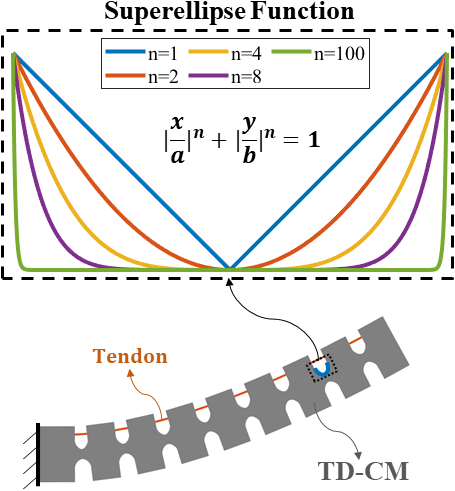}
\caption{Conceptual illustration of a superellipse function (SEF) serving as a geometric tool for design of TD-CMs with discontinuous cross-section geometries. As shown, simply by changing the exponent $n$ in an SEF parametric formulation, different geometric shapes can be assumed for FPs of a TD-CM. }
\label{fig:TDCM-concept}
\end{figure}

	\IEEEPARstart{T}{endon}-Driven Continuum Manipulators (TD-CMs) are  designed to imitate the flexibility and dexterity of natural biological systems \cite{9707607}. These robots are often made  of a flexible backbone actuated by tendons that extend along the structure's length, allowing for controlling the shape, stiffness, and deformation of the robot \cite{burgner2015continuum}.  As opposed to the conventional robots, TD-CMs lack rigid links and recognizable rotational joints, resulting in infinite degrees of freedom (DoF) and the ability to bend continuously through elastic deformation along their body \cite{burgner2015continuum} (see Fig. \ref{fig:TDCM-concept}).  These unique characteristics allow TD-CMs to offer several benefits over rigid-body robots, such as  inherent compliance, light weight, small form-factor, excellent accessibility, dexterity,  and safe interaction with the environment \cite{burgner2015continuum}.  Such features make them a great candidate for various medical procedures in which a TD-CM can safely navigate through tight spaces for diagnosis or treatment purposes \cite{simaan2018medical}. Examples include different minimally invasive surgeries in orthopedics \cite{alambeigi2019use}, single-incision laparoscopic procedures \cite{choi2014surgical}, and eye surgeries \cite{gupta2017toward}.

 The steerability, flexibility, and bending behavior of TD-CMs directly depend on their material constitution (e.g., Nitinol {\cite{rao2023modeling}} or silicon {\cite{renda20123d}}) and geometrical structure (e.g., their shape {\cite{9707607}}, and  continuous {\cite{renda2014dynamic}} or discontinuous {\cite{gao2016mechanical}} geometry), and number of segments (e.g. {\cite{renda2018discrete}}). 
 Using these parameters, various types of TD-CMs  have been introduced by researchers. Nevertheless,  design and development of this type of manipulators typically  suffer from various challenges that need to be addressed {\cite{9707607}}.
First, a review of the literature shows that  design of  TD-CMs with discontinuous geometry (i.e., including notch or flexure patterns (FPs) along the body of robot) is often presumed (e.g., {\cite{Handed2021RAL}})  or  selected based on  finite element analysis (FEA) (e.g., \cite{goury2018fast}, \cite{adagolodjo2021coupling}) and not a generic   analytical modeling framework to thoroughly study the effect of these patterns on the deformation behavior of the manipulator. 
For example, in \cite{pacheco2021beyond}, {\cite{zeng2021modeling}, and \cite{chitalia2020modeling}}, authors have pre-selected a rectangular FP for their TD-CM whereas in \cite{Handed2021RAL} and \cite{gao2016mechanical}, a   circular pattern has been used and presumed    without  optimally designing this pattern by performing an analytical design procedure. It is worth highlighting that although FEA has been suggested as a  tool for the design and deformation modeling of soft robots  (e.g., \cite{goury2018fast,adagolodjo2021coupling, chitalia2020modeling, childs2022improving}), performing reliable and valid FEA of a TD-CM with discontinuous geometry is often time-consuming and very challenging, especially when considering   large deformation for these robots and interactions between tendons and the robot body causing hysteresis and tension loss due to friction \cite{bieze2018finite,liu2021influence,liuTmech2022}.
% GIVE SOME EXAMPLES HERE OF THE PAPERS HEURISTICALLY SELECTED NOTCHES OR USED FE
% However, this approach can limit the ability to design and achieve the desired level of flexibility and steerability \cite{thomas2020monolithic, gao2016mechanical}.
% Additionally, existing approaches are often developed for or tailored to specific structures and geometries, such as rectangular or circular geometry \cite{thomas2020monolithic, gao2016mechanical}, which limit their applicability for a generic-purpose application.
Therefore, there is a crucial need for a more comprehensive and generic mathematical framework that can readily enable design   of a wide range of TD-CMs with distinct flexural patterns and analyzing their effects on the deformation behavior of TD-CMs.  

 Aside from the effect of FPs  on the deformation behavior of TD-CMs, 
 % although TD-CMs with discontinuous cross-section geometries or flexure patterns hold great potential for various surgical applications, accurately modeling and simulating their behavior present significant challenges (e.g., \cite{liuTmech2022}).
 % modeling  tendons, their routing and  interactions with the  tendons' channels are extremely challenging yet crucially important for optimal design and control of such manipulators. actuation of TD-CMs typically leads to highly-nonlinear effects, such as tension loss and hysteresis that directly impact the deformation behavior of these robots \cite{simaan2018medical, subramani2015tackling, yang2020}.
 this important design parameter also directly affects  the tension loss and hysteresis of these robots. As shown in  \cite{simaan2018medical, subramani2015tackling, yang2020,liu2021influence,liuImpact2022,liuTmech2022}, the deformation behavior  and tension-loss of TD-CMs are inter-related and need to be simultaneously considered during the design and modeling of such manipulators. 
 % Nevertheless, modeling this complex and nonlinear relationship is extremely challenging \cite{yang2020}.
% In our previous studies (i.e., \cite{yang2020} and \cite{liu2021effect}), 
Nevertheless, to simplify this complex and nonlinear problem, most of the existing literature   neglect to consider the concurrent effect of robot geometry and loss of tendons' tensions because of friction on robot deformation behavior. Examples include but not limited to the studies using piecewise constant-curvature  (e.g., \cite{webster2010design}, \cite{camarillo2008mechanics}) and variable-curvature methods (e.g., \cite{renda20123d}, \cite{boyer2020dynamics})  solely focusing on modeling  deformation behavior of TD-CM in an ideal friction-less situation. Although there exist studies that have considered the  effects of friction in modeling deformation behavior of TD-CMs and tendon-sheath systems, they typically rely on simplifying assumptions or models such as models using lumped parameters \cite{kato2016tendon}, pre-defined or known tendon configurations (e.g., tendon wrap angle \cite{sun2013elongation} and tendon curvature distribution \cite{agrawal2010modeling}), or experimentally-tuned  parameters that are not generic and can be applied on particular types of TD-CMs and experimental conditions \cite{roy2016modeling}.

To partially address the above-mentioned limitations,  in \cite{yang2020}, \cite{liu2021effect}, \cite{liu2021influence} and \cite{liuImpact2022}, we proposed  analytical modelling frameworks  that fully describe the unknown correlation between internal forces and deformation behaviors of TD-CMs with continuous geometries without relying on a priori known information or experimentally-tuned parameters. However, similar to the most friction-less and friction-based models in the literature (e.g., \cite{9707607}), our model did not account for the effect of discontinuous cross-section geometries or FPs on TD-CM design and deformation behavior. Particularly, the proposed models, cannot completely describe the effect of  distributed and concentrated point friction forces caused by partially-constrained tendons inside the tendon channels of TD-CMs. 

To  address the aforementioned limitations, and as our main contributions, in this work,
(1) we propose  a unique mathematical framework that
% and solution algorithm  that can be used for  (i) the design  of a  TD-CM with   discontinuous cross-section geometry  and a generic  FP  and (ii) modeling the inter-related effect of tendons' friction and internal forces on the tension loss and deformation behavior of this manipulator. Particularly, our framework
features the \textbf{\textit{superellipse functions}} \cite{gridgeman1970lame} (see Fig. \ref{fig:TDCM-concept})  as a  geometric and versatile tool for effective design and modeling of a generic FP (e.g., triangular, rectangular, or circular)  in TD-CMs with  discontinuous cross-section geometries;   
% study presents significant advancements over our previous works (i.e., \cite{yang2020}, \cite{liu2021effect}, \cite{liu2021influence} and \cite{liuImpact2022}) to address the aforementioned limitations. Firstly, 
% we introduce a novel, generic geometric-modeling-based framework for TD-CMs that can accommodate a wide range of discontinuous cross-section geometries.
% This framework maps geometric parameters to desired performance metrics, facilitating the design and tuning of robotic TD-CMs. We developed this modeling framework based on the \textbf{\textit{superellipse function}} (see Fig. \ref{fig:TDCM-concept}) to serve as a geometric modeling tool that enables effective design and optimization of TD-CMs with discontinuous cross-section geometries or flexure patterns.
(2) We also propose a unique complementary mathematical framework and solution algorithm for  modeling the effects of distributed and concentrated  friction forces--  caused by partially-constrained tendons-- on tension loss and deformation behavior of a TD-CM designed and modeled by a superellipse function. 
Particularly, by employing this geometric function, we establish four intuitive functional  mappings relating the manipulator's geometric parameters, kinematics, and deformation behavior enabling efficient design and optimization of TD-CMs with complex FP geometries;
(3) Thanks to the features of these intuitive mappings and superellipse function, we also introduce the concepts of \textit{design space} and \textit{design curve} to intuitively design a generic TD-CM with FPs experiencing tension loss; and
(4) To thoroughly evaluate the performance of this novel framework and its aforementioned features, 
% Secondly, we propose a novel solution algorithm that can solve a TD-CM model with generic discontinuous cross-section geometries or flexure patterns without requiring the solution of the classic Cosserat rod differential equilibrium equation and its boundary value problem formulation. This is particularly advantageous for TD-CMs with discontinuous geometries, as their second moment of area functions may not be differentiable, making the solution of the Cosserat rod model challenging \cite{rucker2011statics}.
% Thirdly, contrary to existing models, our model can effectively model the effects of both resulting distributed and concentrated point friction caused by partially-constrained tendon routing paths.
 we performed various simulation and experimental studies on TD-CMs designed based on a superellipse function. Results were then analyzed and compared  using different error metrics.

% Although accurate models have been studied for existing structure- (e.g., \cite{thomas2020monolithic}), geometry- (e.g., \cite{wang2021evolutionary}), and design-specific applications (e.g., \cite{hwang2020review}), these methods can not cover \textbf{generic cross-section geometry (e.g., generic notches pattern)}. Therefore, they can not be readily extended to other forms of continuum robot with a continuous or discontinuous  cross section geometry.

% Accurate models are essential for designing TD-CMs, their actuation unit, and predicting the performance of the robot in various scenarios (e.g.,  \cite{simaan2018medical}). A review of the literature demonstrates that  there is still a lack of appropriate analytical modeling tools for the effective design and optimization of TD-CMs with discontinuous cross-section geometries or flexure patterns along their geometry.
 
\section{Mathematical Modeling} 
 The shape and geometry of FPs  play a crucial role in determining the TD-CM's overall bending behavior and workspace. To effectively  design these patterns for a generic TD-CM, in this section, we first feature the Superellipse function \cite{gridgeman1970lame} (SEF) as a geometric and versatile tool to analytically design the geometry and shape of these patterns for a generic TD-CM.
We then derive generic equations to represent  kinematics of such a TD-CM.  Finally, we introduce unique equations representing the interrelated relationship between the  deformation behavior and tension-loss of TD-CMs  and propose an iterative solution algorithm to solve such nonlinear equations.  

% In this section, to enable effective design and optimization of TD-CMs with discontinuous cross-section geometries or flexure patterns, we introduce a novel, generic geometric-modeling-based framework for TD-CMs that can accommodate a wide range of discontinuous cross-section geometries.  This modeling framework is based on the superellipse function and can serve as a geometric modeling tool and map geometric parameters to desired performance metrics.

\begin{figure*}[!t]
\centering
\includegraphics[width=1\linewidth]{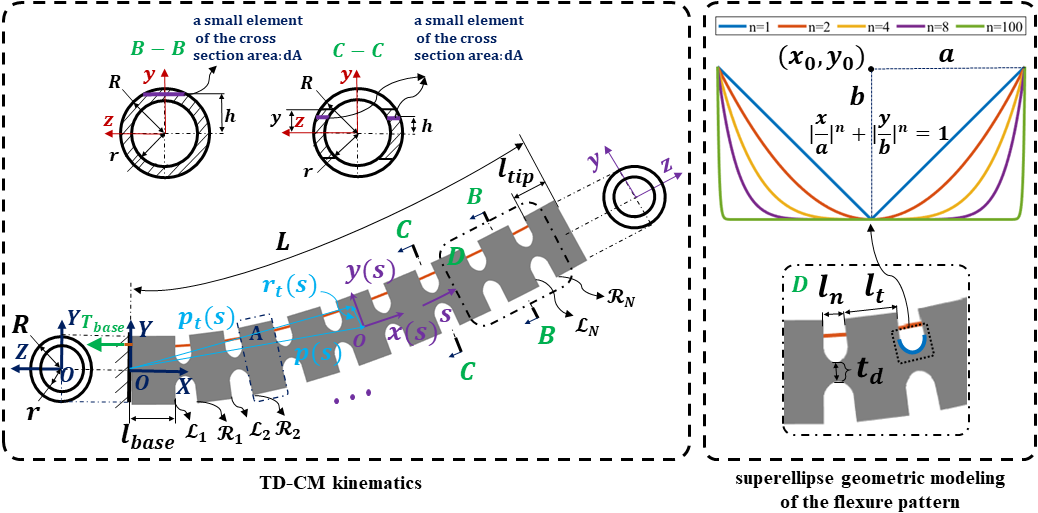}
\caption{Conceptual illustration of  a  TD-CM  with circular cross sectional geometry and $N$ evenly and symmetrically distributed FPs designed based on an SEF. Figure also shows the parameters and reference frames used to model such TD-CMs.}
\label{fig:modeling1}
\end{figure*}

\subsection{The Superellipse Function (SEF)}
\label{sec: superellipse modeling}
 % Discontinuous cross-sectional geometries or flexure patterns  are often used in TD-CMs structures to control their deformation behavior and workspace. 

An SEF, also known as Lamé curve \cite{gridgeman1970lame},  efficiently represents a wide family of geometric curves
% \yl{\sout{that are similar to an ellipse} (SEF's shape is not similar to the shape of ellipse. SEF and ellipse share the same geometric features of semi-major axis and semi-minor axis, and symmetry about them)}
\cite{barr1981superquadrics} by  solely utilizing  three constant parameters.  In a 2-Dimensional (2D)  Cartesian coordinate system,  any set of points, represented by the pair $(x\in\mathbb{R}, y\in\mathbb{R})$ on an SEF curve,   can be characterized by the following implicit function: 
\begin{align}
    \vert\frac{x}{a}\vert^n+\vert\frac{y}{b}\vert^n = 1 \label{eq:superellipse equation1}
\end{align}
where $|..|$ indicates the absolute value of a  number. Parameters $a\in \mathbb{R}^+$ and $b\in \mathbb{R}^+$ are real and positive numbers representing  semi-diameters of the curve and exponent $n\in \mathbb{R}^+$ is a real and positive value determining the overall shape of the curve. 
Fig. \ref{fig:TDCM-concept} represents example curves generated by changing the  exponent $n$ in (\ref{eq:superellipse equation1}). As shown,  $n=1$, $n=2$, and $n=100$ results in a triangular, circular, and rectangular curve, respectively. 

Equation (\ref{eq:superellipse equation1}) can also be expressed in the following  parametric form in which $0 \leq t \leq 2\pi$ is a parameter having no elementary geometric interpretation: 
% (represented by $\mathcal{S}$) 
% can be further expressed using parametric form of superellipse function as:
\begin{align} \label{eq:superellipse equation2}
\begin{cases}
x = x_0 + a |\cos{t}|^{\frac{2}{n}}   \text{sgn}(\cos{t})  \\[8pt]
y = y_0 + b |\sin{t}|^{\frac{2}{n}}   \text{sgn}(\sin{t})
\end{cases}
\end{align}
where $x_0$ and $y_0$ define the origin of the SEF.   Similar to (\ref{eq:superellipse equation1}), by adjusting the value of exponent $n$ in the SEF, the shape of the curve   can be varied (see Fig. \ref{fig:TDCM-concept}). Also, the $sgn$ function is defined as:
\begin{align}
\nonumber\text{sgn}(w)=
\begin{cases}
+1, \hspace{30pt}w>0\\
0,  \hspace{38pt}w=0\\
-1, \hspace{30pt}w<0
\end{cases}
\end{align}
% to achieve the desired bending characteristics. 

Considering (\ref{eq:superellipse equation1}) and (\ref{eq:superellipse equation2}), the main advantage of utilizing an SEF, therefore, is its ability to construct a rich group of dis/continuous curves by solely modifying three  parameters \cite{barr1981superquadrics}. This feature makes the SEF an ideal and versatile tool for generating various FPs  for designing a generic TD-CM (see Fig. \ref{fig:modeling1}). In other words, by adjusting the parameters of an SEF (i.e., $a$, $b$, and $n$), one  can simply define  the shape of the FP  to achieve a specific deformation behavior and workspace for a TD-CM.

\subsection{TD-CM Geometric Modeling}

As shown in Fig. \ref{fig:modeling1} and without loss of generality, let's consider a  TD-CM design with circular cross sectional geometry and $N$ evenly and symmetrically distributed FPs  on both sides of the robot's body. Of note, for this TD-CM,  by changing the exponent $n$ in (\ref{eq:superellipse equation1}), the geometry of FPs can readily be modified and designed to obtain the required deformation behavior and workspace (see Fig. \ref{fig:modeling1}).  The overall length $L$ of such TD-CM can be obtained by the following expression:
\begin{align} 
\label{eq:overall length}
L = l_{base} + N l_n + (N-1) l_t+ l_{tip}
\end{align}
where, as shown in Fig. \ref{fig:modeling1}, $l_{base}$ and $l_{tip}$ represent the lengths of the proximal and distal segments of the TD-CM, respectively. $l_n$ denotes the width of each FP, $t_d$ indicates the width of the TD-CM's backbone, and the distance between two adjacent FP is determined by $l_t$.  Of note, besides adjusting the value of exponent $n$  in (\ref{eq:superellipse equation1}), these parameters also directly determine the deformation behavior and workspace of a TD-CM  (see Fig. \ref{fig:modeling1}) .

% where, $l_{base}$ and $l_{tip}$ denote the lengths of the proximal and distal segments of the TD-CM, respectively. $l_n$ represents the width of each notch, while $l_t$ denotes the distance between two adjacent notches. The width of the TD-CM's backbone is represented by $w_d$, as shown in Fig. xx. Importantly, by adjusting the value of exponent $n$ in the superellipse function and these aforementioned design parameters, the flexibility and workspace of the TD-CM can be customized and tailored to meet specific application requirements.

We can   define the domain of the notched sections $\mathcal{D}_{notched}$ (i.e.,  sections across the TD-CM  with an FP)  and the tubular sections $\mathcal{D}_{tubular}$ (i.e.,  sections across the TD-CM without an FP)   of the TD-CM  as follows: 
% $D_{notched}=\{s: l_i\leq\ s \leq r_i\}$ and $D_{tubular}=\{s: r_{i}\leq\ s \leq l_{i+1}\}$.
\begin{align} \label{eq:domain}
\begin{cases}
\mathcal{D}_{notched}=\{s: \mathcal{L}_i< s < \mathcal{R}_i\} \\[8pt]
\mathcal{D}_{tubular}=\{s: \mathcal{R}_{i-1}\leq\ s \leq \mathcal{L}_{i}\}
\end{cases}
\end{align}
where $s$ represents the midline arc length parameter.

Also, the $i^{th}= \{1, 2, ... , N\}$  left boundary $\mathcal{L}_i$ and right boundary  $\mathcal{R}_i$ of the FPs can be expressed as:
\begin{align} \label{eq:boundary}
\begin{cases}
\mathcal{L}_i = l_{base} + (i-1) l_n + (i-1) l_t  \\[8pt]
\mathcal{R}_i = l_{base} + i l_n + (i-1) l_t
\end{cases}
\end{align}

As illustrated in Fig. \ref{fig:modeling1},    cross sectional geometry of the considered TD-CM is discontinuous and varies along its arc length. Section B -- B and Section Section C -- C in the figure clearly represent these changes at the tubular and notched sections, respectively. 
% The notched section possesses a distinct and smaller stiffness compared to the tubular section.
To accurately model the deformation behavior of the TD-CMs, therefore,  it is essential to calculate the effect of such variations on the cross sectional area and   the second moment of area $I(s)$ about the bending axis of the TD-CM  (i.e., the $z$ axis shown in Fig. \ref{fig:modeling1}). 
% Of note,  These calculations are necessary to determine the stiffness of TD-CMs' cross sections, enabling an accurate representation of their deformation behavior.
The following equations represent variation of $I(s)$ across arc length of the TD-CM and, specifically,  at both  notched $\mathcal{D}_{notched}$ and tubular $\mathcal{D}_{tubular}$ domains:
% \begin{align}
% \label{eq:2nd moment of area}
% I(s)=\iint_{\mathcal{S}} h^2dA=
% \begin{cases}
% \displaystyle\frac{\pi}{4}(R^4-r^4)   \hspace{55pt} s  \in \mathcal{D}_{tubular}\\[5pt]
% 2\displaystyle\int_{-y}^{y}h^2\sqrt{R^2-h^2}dh \ldots\\-2\displaystyle\int_{-y}^{y}h^2\sqrt{r^2-h^2}dh
% \hspace{10pt} s  \in \mathcal{D}_{notched}
% \end{cases}
% \end{align}
\begin{align}
\label{eq:2nd moment of area}
I(s)=
\begin{cases}
\displaystyle\frac{\pi}{4}(R^4-r^4)   \hspace{55pt} s  \in \mathcal{D}_{tubular}\\[5pt]
2\displaystyle\int_{-y}^{y}h^2\sqrt{R^2-h^2}dh \ldots\\-2\displaystyle\int_{-y}^{y}h^2\sqrt{r^2-h^2}dh
\hspace{10pt} s  \in \mathcal{D}_{notched}
\end{cases}
\end{align}
% \begin{align} \label{eq: 2nd moment of area-tubular section} 
% I_{tubular}(s)=\iint_{\mathcal{S}} h^2dA
% =\frac{\pi}{4}(R^4-r^4) 
% \end{align}
% For the notched section, the calculation method is as follows:
% \begin{align} \label{eq: 2nd moment of area-notched section} 
% I_{notched}(s)
% &=\nonumber\iint_{\mathcal{S}} h^2dA \\
% &=2\int_{-y}^{y}h^2\sqrt{R^2-h^2}dh-2\int_{-y}^{y}h^2\sqrt{r^2-h^2}dh
% \end{align}
where, as demonstrated in Fig. \ref{fig:modeling1}, $R$ indicates the outer and $r$ represents the inner radii of the TD-CM's circular cross sectional geometry. Moreover, 
% $\mathcal{S}$ is the aforementioned shape or geometry of TD-CM notches curve, 
% $dA$ is the infinitesimal area element, and
$h$ is the perpendicular distance from the axis $z$ to a small element of the cross section area $dA$.
% We also refer the reviewer to reference \cite{agaranamathematical} for more details for calculating the second moment of area.

\begin{remark}
\label{remark1} 
\textit{In this study, without loss of generality,  we are developing equations for a TD-CM with a circular cross sectional geometry and arbitrary FPs. Nevertheless, thanks to the use of SEF, the proposed modeling framework is  generic and can be used for the design and modeling of a TD-CM with a fixed or variable cross sectional geometry. The change in cross sectional geometry will solely affect the calculation of the second moment of area $I(s)$ in (\ref{eq:2nd moment of area}).}
\end{remark}
% In this study, we have considered a circular cross sectional geometry a a wide range of discontinuous cross-section geometries. Specifically, the framework utilizes the superellipse function to enable the mapping of geometric parameters to desired performance metrics, thus serving as an effective geometric modeling tool for TD-CMs.}

\subsection{TD-CM Kinematics Modeling}
% As depicted in Fig. xx, and consistent with the reference frames used in our previous works \cite{liu2021influence}, \cite{liuTmech2022}, \cite{liu2021effect}, we describe the planar deformation behavior of the TD-CM with discontinuous cross-section geometries (i.e., discrete flexural patterns) using the body frame pose along the midline arc length parameter $s\in [0, L]$ with respect to the global frame \{$XYZ$\}. The homogeneous transformation matrix $T(s)\in SE(3)$ is used to represent the transformation from the body frame \{$xyz$\} to the global frame \{$XYZ$\}. Here, $L$ represents the overall length of the TD-CM's midline, $\boldsymbol{p}(s)\in \mathbb{R}^3$ denotes the position vector, and $\mathcal{R}(s)=[\boldsymbol{x}(s),~ \boldsymbol{y}(s),~ \boldsymbol{z}(s)]\in SO(3)$ represents the orientation matrix of the body frame \{$xyz$\}. The rigid-body transformation matrix is expressed as:
Utilizing the body  \{$xyz$\} and global \{$XYZ$\} reference frames, shown in Fig. \ref{fig:modeling1},
% and consistent with the reference frames used in our previous works \cite{liuTmech2022}, \cite{liu2021influence}, \cite{liu2021effect},
we can describe the planar deformation behavior of the TD-CM using the following homogeneous transformation matrix $\mathcal{T}(s) \in SE(3)$:
% ,
% which has discontinuous cross-section geometries (i.e., discrete flexural patterns), 
% using the body frame \{$xyz$\} pose along the midline arc length parameter $s$ in the range of $[0, L]$, with respect to the global frame \{$XYZ$\}.
% The transformation from the body frame \{$xyz$\} to the global frame \{$XYZ$\} is represented by the homogeneous transformation matrix $\mathcal{T}(s) \in SE(3)$.
\begin{align} \label{eq:rigid-body transformation matrix}
\mathcal{T}(s)=\left(
\begin{array}{cc}
\mathcal{R}(s)& \boldsymbol{p}(s) \\[5pt]
\mathbf{0}^T&1 \
\end{array}
\right)
\end{align}
where  $\boldsymbol{p}(s)\in\mathbb{R}^3$ denotes the position vector defining any points on the TD-CM's midline arc length with respect to the global frame \{$XYZ$\}, and $\mathcal{R}(s)=[\boldsymbol{x}(s),~ \boldsymbol{y}(s),~ \boldsymbol{z}(s)]\in SO(3)$ represents the orientation matrix of the body frame \{$xyz$\} with respect to the global frame \{$XYZ$\}. 
% Here, $L$ represents the overall length of the TD-CM's midline, $\boldsymbol{p}(s)\in \mathbb{R}^3$ denotes the position vector, and $\mathcal{R}(s)=[\boldsymbol{x}(s),~ \boldsymbol{y}(s),~ \boldsymbol{z}(s)]\in SO(3)$ represents the orientation matrix of the body frame \{$xyz$\}. The rigid-body transformation matrix is expressed as follows:

The body frame configuration along the arc length parameter $s$ evolves continuously and can be described using the twist vector $V_b(s) = [\omega_b(s), v_b(s)] ^T\in \mathbb{R}^6$. This vector captures the angular $\omega_b(s)\in \mathbb{R}^3$ and linear $v_b(s)\in \mathbb{R}^3$  rate of changes of the configuration $\mathcal{T}(s)$ with respect to $s$. The derivative of $\mathcal{T}(s)$ with respect to $s$, denoted as $\mathcal{T}'(s)$, can  be expressed as:
% The continuous evolution of the body frame configuration along the arc length parameter $s$ is described using the twist vector $V_b(s) = [\omega_b(s), v_b(s)] ^T\in \mathbb{R}^6$, which captures the angular and linear rate of changes of configuration $T(s)$ with respect to $s$. The derivative of $T(s)$ with respect to $s$ (i.e., $T'(s)$) is represented as:
\begin{align} \label{eq:derivative of T}
\mathcal{T}'(s)=
\left(
\begin{array}{cc}
\mathcal{R}(s)& \boldsymbol{p}(s) \\[5pt]
\mathbf{0}^T&1 \
\end{array}
\right)\left(
\begin{array}{cc}
[\omega_b(s)] & v_b(s) \\[5pt]
\mathbf{0}^T & 0 \
\end{array}
\right)
\end{align}

where  $[\omega_b(s)]\in so(3)$ is the skew-symmetric matrix representation of the angular rate of change $\omega_b(s)\in \mathbb{R}^3$. Of note, from now on, the notation `` $'$ '' and `` $''$ '' represent the first and second derivative with respect to $s$, respectively.

% \begin{remark}
% \label{remark2} 
% \textit{The proposed mathematical framework is derived based on Cosserat rod kinematics \cite{NonlinearElasticity} and is more generic. This model can overcome the shortcomings of the existing structure- (e.g., \cite{thomas2020monolithic}), geometry- (e.g., \cite{wang2021evolutionary}), and design-specific models (e.g., \cite{hwang2020review}) and  can be applied for the design of TD-CMs with a continuous or discontinuous  cross section geometry (i.e., with discrete flexural patterns). 
% }
% \end{remark}

\begin{figure*}[!t]
\centering
\includegraphics[width=1\linewidth]{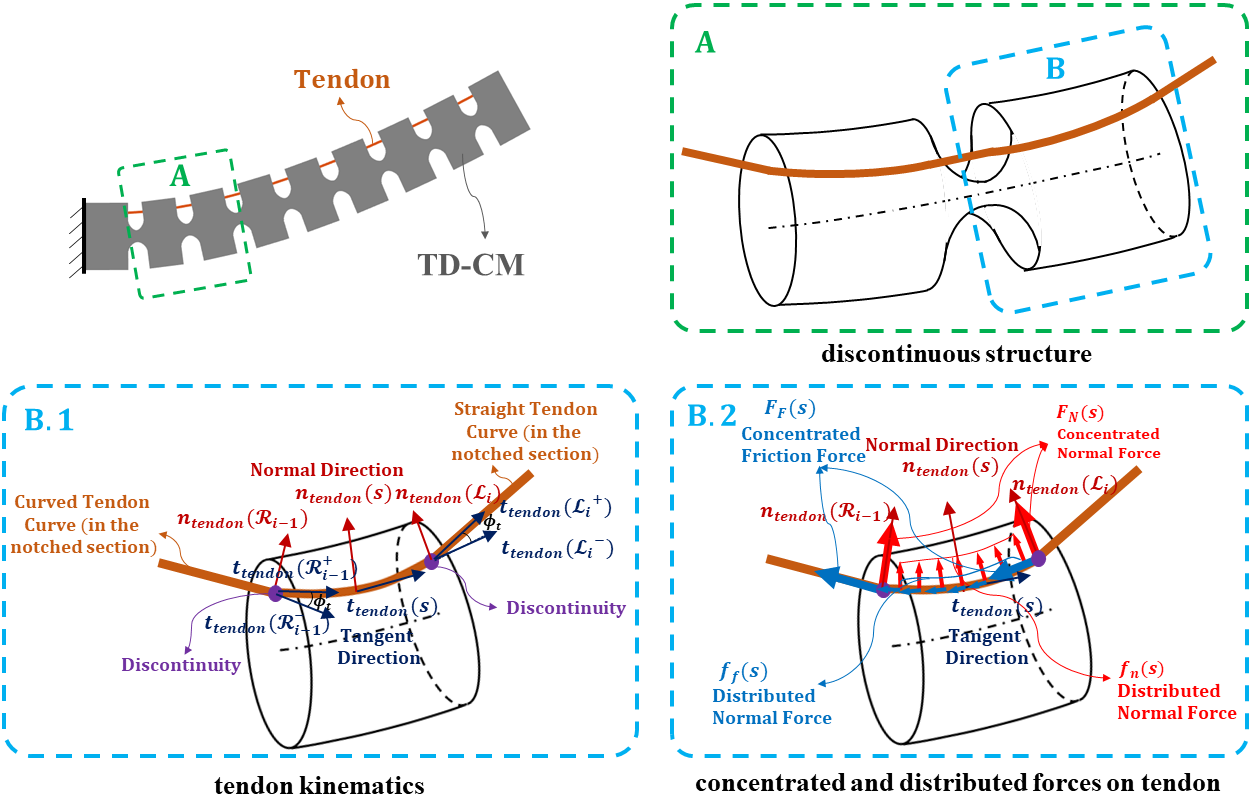}
\caption{Figure shows the internal forces acting on the tendon within the $\mathcal{D}_{notched}$ and  $\mathcal{D}_{tubular}$ domains including the tendon tension $\boldsymbol{T}(s)$, distributed normal force $\boldsymbol{f}_n(s)$, and distributed friction force $\boldsymbol{f}_f(s)$), as well as the concentrated normal force $\boldsymbol{F}_{N}(s)$, and concentrated friction force $\boldsymbol{F}_{F}(s)$ at the exit and entry points from one domain to the other one.}
\label{fig:modeling2}
\end{figure*}

\subsection{Tendons' Kinematics and Statics Modeling}
To completely model the effect of tension-loss on deformation behavior of a TD-CM with discontinuous geometry, we need to completely consider and model all the forces acting on the tendons passing through such manipulators. As shown in Fig.  \ref{fig:modeling1}, for a TD-CM with FPs,  as opposed to the literature modeling a TD-CM with fully constrained tendons (e.g., {\cite{subramani2015tackling}}, \cite{liuTmech2022}, \cite{liu2021influence}, \cite{liu2021effect}), not only we need to model the common internal forces acting on the tendon within the $\mathcal{D}_{notched}$ and  $\mathcal{D}_{tubular}$ domains (i.e., tendon tension $\boldsymbol{T}(s)$, distributed normal force $\boldsymbol{f}_n(s)$, distributed friction force $\boldsymbol{f}_f(s)$), we also should consider and model the concentrated normal force $\boldsymbol{F}_{N}(s)$, and concentrated friction force $\boldsymbol{F}_{F}(s)$ at the exit and entry points from one domain to the other one (see Fig. \ref{fig:modeling2}). 
% To accurately model the forces acting on a tendon passing through  including the tendon tension $\boldsymbol{T}(s)$, distributed normal force $\boldsymbol{f}_n(s)$, distributed friction force $\boldsymbol{f}_f(s)$, concentrated normal force $\boldsymbol{F}_{n}(s)$, and concentrated friction force $\boldsymbol{F}_{f}(s)$., it is necessary to determine the directions of various components, including the tendon tension $\boldsymbol{T}(s)$, distributed normal force $\boldsymbol{f}_n(s)$, distributed friction force $\boldsymbol{f}_f(s)$, concentrated normal force $\boldsymbol{F}_{n}(s)$, and concentrated friction force $\boldsymbol{F}_{f}(s)$. 
To accomplish this, in this section, a thorough derivation of the kinematics of a partially constrained tendon is first presented. These equation will be used in the next section to completely describe the mentioned distributed and concentrated forces acting on the tendon.

% As shown in Fig. \ref{fig:modeling1}, to model forces or loads on a tendon, the directions associated with the tendon tension  $\boldsymbol{T}(s)$, distributed normal force $\boldsymbol{f}_n$, distributed friction $\boldsymbol{f}_f$, concentrated normal force $\boldsymbol{F}_{n}$ and concentrated friction force $\boldsymbol{F}_{f}$ need to be obtained.

% In order to describe the distributed forces and concentrated forces on the tendon, the kinematics of a tendon partially constrained is thoroughly derived in this section.
\subsubsection{Tendon Kinematics Tuple Function}
To define the tendon's kinematics, we first need to  calculate a position vector  $\boldsymbol{p}_{t}(s)$ with respect to the global frame \{$XYZ$\} (see Fig. \ref{fig:modeling1}) that defines the  location of tendon inside the constrained    (i.e., parts of the tendon located within the  tubular domain $\mathcal{D}_{tubular}$)  or unconstrained  (located within the notched domain $\mathcal{D}_{notched}$) segments of the TD-CM as follows:
% \begin{align} \label{eq:psotion vector of tendon1} 
% \boldsymbol{p}_{t}(s)=\boldsymbol{p}(s)+\boldsymbol{r}_{t}(s)  \hspace{20pt} s  \in \mathcal{D}_{tubular}
% \end{align}
\begin{align}
\label{eq:psotion vector of tendon}
\boldsymbol{p}_{t}(s)=
\begin{cases}
\boldsymbol{p}(s)+\boldsymbol{r}_{t}(s)  \hspace{78pt} s  \in \mathcal{D}_{tubular}\\[5pt]
\boldsymbol{p}(\mathcal{R}_{i-1})+\boldsymbol{r}_{t}(\mathcal{R}_{i-1}) + [\boldsymbol{p}(\mathcal{L}_{i})+\boldsymbol{r}_{t}(\mathcal{L}_{i})-\boldsymbol{p}(\mathcal{R}_{i-1})\\-\boldsymbol{r}_{t}(\mathcal{R}_{i-1})]  \displaystyle\frac{s-\mathcal{R}_{i-1}}{\mathcal{L}_{i}-\mathcal{R}_{i-1}}
\hspace{25pt} s  \in \mathcal{D}_{notched}
\end{cases}
\end{align}
where, as shown in  Fig. \ref{fig:modeling1}, 
% $\boldsymbol{p}(s)$ denotes the position vector of TD-CM's midline and
$ \boldsymbol{r}_{t}(s)$ denotes a  vector from  origin of the body frame \{$xyz$\} to the corresponding point on the tendon. 
% The tendon located within the notched domain $\mathcal{D}_{notched}$ is expressed using linear function (i.e.,  a  straight line).

An important point in calculating the kinematics of a partially constrained tendon  in TD-CM with FPs is that its shape and particularly curvature dramatically changes in tubular or notched domains. As conceptually illustrated in  Fig. \ref{fig:modeling2}, in tubular domains $\mathcal{D}_{tubular}$, tendon's shape adapts the TD-CM's curvature whereas it would have  zero curvature (i.e.,  a  straight line) in unconstrained notched domains $\mathcal{D}_{notched}$ .  
% curved during TD-CM's deformation. In contrast, the tendon located in the notched section without any constraints remains a straight line.
To calculate the tendon's curvature, as shown in Fig. \ref{fig:modeling2}, we first calculate the unit tangent   $\boldsymbol{t}_{\text{tendon}}(s)$ and normal $\boldsymbol{n}_{\text{tendon}}(s)$ vectors of the tendon using the Frenet–Serret formulas \cite{mate2017frenet}, \cite{liuImpact2022}:
\begin{align}
\boldsymbol{t}_{\text{tendon}}(s)=
\begin{cases}
\displaystyle\frac{\boldsymbol{p}'_{t}(s)}{\left\Vert \boldsymbol{p}'_{t}(s)\right\Vert} \hspace{72pt} s  \in \mathcal{D}_{tubular}\\[10pt]
\displaystyle\frac{\boldsymbol{p}_{t}(\mathcal{L}_{i+1})-\boldsymbol{p}_{t}(\mathcal{R}_i)}{\left\Vert \boldsymbol{p}_{t}(\mathcal{L}_{i+1})-\boldsymbol{p}_{t}(\mathcal{R}_i)\right\Vert} \hspace{15pt} s  \in \mathcal{D}_{notched}
\end{cases}
\label{eq:tangent direction}
\end{align}

\begin{align} 
 \boldsymbol{n}_{\text{tendon}}(s)=
\begin{cases}
    \displaystyle\frac{ \boldsymbol{p}'_{t}(s)\times(\boldsymbol{p}''_{t}(s)\times\boldsymbol{p}'_{t}(s))}{\left\Vert\boldsymbol{p}''_{t}(s)\times\boldsymbol{p}'_{t}(s)\right\Vert\left\Vert\boldsymbol{p}'_{t}(s)\right\Vert}
    \hspace{10pt} s  \in \mathcal{D}_{tubular}\\[10pt]
    \boldsymbol{0} \hspace{115pt} s  \in \mathcal{D}_{notched}
\end{cases}
\label{eq:normal direction}
\end{align}
 where $||..||$ indicates the two-norm of a vector. Also, ``$\times$'' represents the cross  products of two vectors, respectively.

% where $\boldsymbol{p}(s)$ denotes the position vector of TD-CM's midline and $ \boldsymbol{R}_{c}(s)$ denotes a  vector from  origin of the body frame \{$xyz$\} to the corresponding point on the tendon.

Using (\ref{eq:tangent direction}) and (\ref{eq:normal direction}), we can obtain the  tendon's curvature $\kappa_t(s)$ in either of the tubular or notched domains as follows:
\begin{align}
\kappa_t(s)=
\begin{cases}
    \displaystyle\frac{\left\Vert \boldsymbol{p}'_{t}(s)\times\boldsymbol{p}''_{t}(s)\right\Vert}{\left\Vert \boldsymbol{p}'_{t}(t)\right\Vert^3}\hspace{25pt} s  \in \mathcal{D}_{tubular}\\[10pt]
    0 \hspace{90pt} s  \in \mathcal{D}_{notched}
\end{cases}
\label{eq:curvature}
\end{align}

Tendon wrap angle   can also be calculated using  the tendon's curvature $\kappa_t(s)$ \cite{liuImpact2022}:
\begin{align}
    \displaystyle\theta_t(s)=\int_0^s \left\Vert \boldsymbol{p}'_{t}(\xi)\right\Vert\kappa_{t} (\xi) \, d\xi
\label{eq:wrap_angle2}
\end{align}

In the modeling process, it is also important to  consider the discontinuous boundary points at which a tubular domain meets a notched domain (see Fig. \ref{fig:modeling2}). At these  exit or entry boundary points from one domain to the other one, concentrated normal force $\boldsymbol{F}_{N}(s)$ and concentrated friction force $\boldsymbol{F}_{F}(s)$   must be taken into account and calculated. To accurately model these forces, as shown in Fig. \ref{fig:modeling2},  we need to first calculate  the angle $\phi_t$-- defining the change  in the tangent direction of the tendon $\boldsymbol{t}_{\text{tendon}}$ when it enters/exits to/from the  tubular domain-- as follows: 
% change in the  the angles $\phi_t$ in the tangent direction of the tendon curve at these points of discontinuity can be expressed as follows:
\begin{align} \label{eq: change in angle}
\phi_t(s) =
    \begin{cases}
         \cos^{-1}\displaystyle\frac{\boldsymbol{t}_{\text{tendon}}(\mathcal{L}_i^{-})  \boldsymbol{t}_{\text{tendon}}(\mathcal{L}_i^{+}) }{\left\Vert\boldsymbol{t}_{\text{tendon}}(\mathcal{L}_i^{-})\right\Vert  \left\Vert \boldsymbol{t}_{\text{tendon}}(\mathcal{L}_i^{+}) \right\Vert} \hspace{20pt} s \in \mathfrak{L}\\[12pt]
         \cos^{-1}\displaystyle\frac{\boldsymbol{t}_{\text{tendon}}(\mathcal{R}_i^{-}) \boldsymbol{t}_{\text{tendon}}(\mathcal{R}_i^{+}) }{\left\Vert \boldsymbol{t}_{\text{tendon}}(\mathcal{R}_i^{-}) \right\Vert  \left\Vert\boldsymbol{t}_{\text{tendon}}(\mathcal{R}_i^{+})\right\Vert} \hspace{18pt} s \in \mathfrak{R}\\[12pt]         
    \end{cases}
\end{align}
where set $\mathfrak{L}=\{\mathcal{L}_1, \mathcal{L}_2, \ldots, \mathcal{L}_N\}$ and set $\mathfrak{R}=\{\mathcal{R}_1, \mathcal{R}_2, \ldots, \mathcal{R}_N\}$ are calculated based on (\ref{eq:boundary}).
Of note, the symbol ``$-$" is used when $s$ approaches $\mathcal{L}_i$ or $\mathcal{R}_i$ from the left, while the symbol ``$+$" is used when $s$ approaches $\mathcal{L}_i$ and $\mathcal{R}_i$ from the right.

\begin{remark}
\label{remark2} 
\textit{
% The five kinematics functions  (\ref{eq:psotion vector of tendon})-(\ref{eq: change in angle}) play a crucial role in characterizing the kinematics of a partially constrained tendon. These functions are utilized to reconstruct the complete state of forces acting on the tendon as well as the distributed wrenches along the TD-CM's midline. Together, 
The aforementioned 
collection of five kinematics functions (\ref{eq:psotion vector of tendon})-(\ref{eq: change in angle}) constitutes the tendon kinematics tuple function denoted as $(\boldsymbol{p}_{t}, \boldsymbol{t}_{\text{tendon}}, \boldsymbol{n}_{\text{tendon}},  \kappa_t, \phi_t )$ -- that serves as the foundational components for determining internal forces in the subsequent sections.
}
% \yl{The aforementioned five kinematics functions (\ref{eq:psotion vector of tendon})-(\ref{eq: change in angle}) 
% play an important role in describing
% kinematics of a partially constrained tendon and will be used to reconstruct the full state of forces on the tendon and distributed wrenches on the TD-CM's midline. The aforementioned collection of five kinematics functions (\ref{eq:psotion vector of tendon})-(\ref{eq: change in angle}) form a tendon kinematics tuple function $(\boldsymbol{p}_{t}, 
% \boldsymbol{t}_{\text{tendon}}, \boldsymbol{n}_{\text{tendon}},  \kappa_t, \phi_t )$ which will be the building blocks to build internal forces in the following sections.}
% In the next section,  we introduce the tendon kinematics tuple function $\boldsymbol{T}(\boldsymbol{p}_{t}, 
% \boldsymbol{t}_{\text{tendon}}, \boldsymbol{n}_{\text{tendon}},  \kappa_t, \phi_t )$
% which is a function defined based on the aforementioned collection of five kinematics functions (\ref{eq:psotion vector of tendon})-(\ref{eq: change in angle}) and will be used to reconstruct the full state of forces on the tendon and distributed wrenches on the TD-CM's midline.}
\end{remark}

\subsubsection{Tendon Statics Tuple Function}
% \subsubsection{Tendon Tension Distribution}
When a tendon is partially constrained within a TD-CM with 
discontinuous cross sectional geometry, unlike the case of a  fully constrained tendon, it experiences both the exponential distributed  and concentrated forces that together cause tension loss \cite{liuImpact2022} . To completely model these forces and their impacts on the tension loss, in this section, we first introduce a force tuple function and use that to derive a generalized and physically-intuitive formulation to express the tendon's statics.

\textbf{Tendon Tension Distribution:}
As shown in Fig. \ref{fig:modeling2} and using the tendon kinematics tuple function $(\boldsymbol{p}_{t}, 
\boldsymbol{t}_{\text{tendon}}, \boldsymbol{n}_{\text{tendon}},  \kappa_t, \phi_t )$ mentioned in Remark \ref{remark2}, tendon tension $\boldsymbol{T}(s)$ along the TD-CM
can be expressed as \cite{liuImpact2022}:
% arc length parameter $s$ is dependent on the tendon wrap angle $\theta_t(s)$ (as shown in Fig. \ref{fig:modeling1}), which can be expressed using the 5-tuple function basis $\langle \boldsymbol{p}_{t}(s), \boldsymbol{t}_{\text{tendon}}(s), \boldsymbol{n}_{\text{tendon}}(s),  \kappa_t(s), \phi_t(s)\rangle$ of tendon kinematics as follows:
% \begin{align} 
% \label{eq:capstan equation}
% \boldsymbol{T}(s)
% &=\nonumber- \boldsymbol{t}_\text{tendon}(s) T(\xi)  e^{-\mu  (\theta_t(s)-\theta_t(\xi)) }\\
% &=-\boldsymbol{t}_\text{tendon}(s) T(\xi) \underbrace{e^{-\lambda\mu  \int_0^s \left\Vert \boldsymbol{p}'_{c}(\xi)\right\Vert\kappa_{c} (\xi) \, d\xi }}_{\boldsymbol{\mathcal{F^{\star}}_{t.l.}}}
% \end{align}

% \begin{align}
% \boldsymbol{T}(s)=
%     \nonumber- \boldsymbol{t}_\text{tendon}(s) T(\xi)  e^{-\mu  (\theta_t(s)-\theta_t(\xi)) }  \\[10pt]
%     \xi \in \mathfrak{R}, s  \in \mathcal{D}_{tubular}
% \label{eq:curvature}
% \end{align}
\begin{align}
\boldsymbol{T}(s)=
\begin{cases}
    - \boldsymbol{t}_\text{tendon}(s) T(\mathcal{R}_i^{+})  \underbrace{e^{-\mu  (\theta_t(s)-\theta_t(\mathcal{R}_i))}}_{\boldsymbol{\mathcal{G}(s)}}  \hspace{3pt} s  \in \mathcal{D}_{tubular}\\[10pt]
     - \boldsymbol{t}_\text{tendon}(s) T(\mathcal{L}_i^{+})  \hspace{75pt} s \in \mathcal{D}_{notched}
\end{cases}
\label{eq:tendon tension}
\end{align}
where,
$\mu$     indicates the frictioncoefficient. Note that $\boldsymbol{T}(s=0)$ shows the input tendon tension $\boldsymbol{T}_{\text{base}}$. 

 In (\ref{eq:tendon tension}),  the term $\boldsymbol{\mathcal{G}(s)}$ represents   the ``\textit{tension loss factor}".  Using (\ref{eq:psotion vector of tendon})-(\ref{eq:wrap_angle}), this term $\boldsymbol{\mathcal{G}(s)}$ can also be  reformulated based on tendon's curvature $\kappa_t(s)$ and position vector  $\boldsymbol{p}_{t}(s)$:
\begin{align} 
\label{eq:wrap_angle}
\boldsymbol{\mathcal{G}(s)}
=e^{-\mu  (\int_0^s \left\Vert \boldsymbol{p}'_{t}(\xi)\right\Vert\kappa_{t} (\xi) \, d\xi - \int_0^{\mathcal{R}_i} \left\Vert \boldsymbol{p}'_{t}(\xi)\right\Vert\kappa_{t} (\xi) \, d\xi )}
\end{align}

Also in (\ref{eq:tendon tension}), $ T(\mathcal{L}_i^{+})$ and  $T(\mathcal{R}_i^{+})$ represent the tendon-tension boundary conditions at the $i$-th left boundary $\mathcal{L}_i$ and right boundary  $\mathcal{R}_i$ of the notched domains, respectively,  and can be defined as follows:
\begin{align}
\begin{cases}
    T(\mathcal{R}_i^{+}) = T(\mathcal{R}_i^{-})-F_F(\mathcal{R}_i) \hspace{30pt} s  \in \mathcal{D}_{tubular}\\[10pt]
     T(\mathcal{L}_i^{+}) = T(\mathcal{L}_i^{-})-F_F(\mathcal{L}_i) \hspace{35pt} s \in \mathcal{D}_{notched}
\end{cases}
\label{eq:boundry conditions}
\end{align}
where $\boldsymbol{F}_F(\mathcal{L}_i)$ and  $\boldsymbol{F}_F(\mathcal{R}_i)$ are the concentrated friction  forces at the $i$-th left  $\mathcal{L}_i$ and right   $\mathcal{R}_i$ boundaries of the notched domains, respectively (see Fig. \ref{fig:modeling2}). Check (\ref{eq:concentrated frictional force}) for definition of $\boldsymbol{F}_F$.
% can be calculated using this equation   \yl{$\boldsymbol{T}(s=0)$ is defined as $\boldsymbol{T}_{\text{base}}$ which is the input tension.} 
% $\boldsymbol{T}(s=0)=\boldsymbol{T}_{\text{base}}$ is the input tension. 
% Due to the propagation of tendon tension along the tendon channel, the calculation of tendon tension is performed in a sequential manner, starting from the proximal end  of the TD-CM and progressing towards its distal end.
% In the exponential term of (\ref{eq:tendon tension}), $\mu$ is frictional coefficient and.

% The tendon-tension boundary conditions at the $i$-th left boundary $\mathcal{L}_i$ and right boundary  $\mathcal{R}_i$ of notched sections are defined as follows:
% \begin{align}
% \begin{cases}
%     T(\mathcal{R}_i^{+}) = T(\mathcal{R}_i^{-})-F_F(\mathcal{R}_i) \hspace{30pt} s  \in \mathcal{D}_{tubular}\\[10pt]
%      T(\mathcal{L}_i^{+}) = T(\mathcal{L}_i^{-})-F_F(\mathcal{L}_i) \hspace{35pt} s \in \mathcal{D}_{notched}
% \end{cases}
% \label{eq:boundry conditions}
% \end{align}

\textbf{Distributed Normal and Friction Forces: 
} As shown in Fig. \ref{fig:modeling2}, when a tendon passing  through the  $\mathcal{D}_{tubular}$  domains, aside from the tendon tension $\boldsymbol{T}(s)$,  it also experiences the distributed normal  $\boldsymbol{f}_n(s)$ and distributed friction forces $\boldsymbol{f}_f(s)$.   As shown in the zoomed region B of Fig.    \ref{fig:modeling2} and  using the tendon kinematics tuple function $(\boldsymbol{p}_{t}, 
\boldsymbol{t}_{\text{tendon}}, \boldsymbol{n}_{\text{tendon}},  \kappa_t, \phi_t )$, distributed normal force $\boldsymbol{f}_n(s)$ and the distributed friction force $\boldsymbol{f}_f(s)$ on the tendon can be expressed as follows:
% using the 5-tuple function basis $\langle \boldsymbol{p}_{t}(s), \boldsymbol{t}_{\text{tendon}}(s), \boldsymbol{n}_{\text{tendon}}(s),  \kappa_t(s), \phi_t(s)\rangle$ of tendon kinematics as:
\begin{align}
\boldsymbol{f}_n(s)=
\begin{cases}
    \underbrace{\boldsymbol{n}_{\text{tendon}}(s) \kappa_t(s)}_{\boldsymbol{\mathcal{P}_n(s)}} T(\mathcal{R}_i^{+})  \underbrace{e^{-\mu  (\theta_t(s)-\theta_t(\mathcal{R}_i))}}_{\boldsymbol{\mathcal{G}(s)}} \\ \hspace{120pt} s  \in \mathcal{D}_{tubular}\\
     \boldsymbol{0}  \hspace{115pt} s \in \mathcal{D}_{notched}
\end{cases}
\label{eq:distributed normal force}
\end{align}
% Moreover, the distributed friction force $\boldsymbol{f}_f(s)$ on the tendon can also be derived  as:
\begin{align}
\boldsymbol{f}_f(s)=
\begin{cases}
    \underbrace{-\mu \boldsymbol{t}_{\text{tendon}}(s) \kappa_t(s)}_{\boldsymbol{\mathcal{P}_f(s)}} T(\mathcal{R}_i^{+})  \underbrace{e^{-\mu  (\theta_t(s)-\theta_t(\mathcal{R}_i))}}_{\boldsymbol{\mathcal{G}(s)}} \\ \hspace{135pt} s  \in \mathcal{D}_{tubular}\\
     \boldsymbol{0}  \hspace{130pt} s \in \mathcal{D}_{notched}
\end{cases}
\label{eq:distributed friction force}
\end{align}
where the terms $\boldsymbol{\mathcal{P}_n}$ and $\boldsymbol{\mathcal{P}_f}$ indicate the ``\textit{projection factors of distributed forces}" that project the tendon tension $\boldsymbol{T}(s)$ onto the distributed normal   $\boldsymbol{f}_n(s)$ and the the distributed friction  $\boldsymbol{f}_f(s)$ forces, respectively.
Detailed derivation of (\ref{eq:distributed normal force}) and (\ref{eq:distributed friction force}) can be found in Appendix \ref{derivation1}.

\textbf{Concentrated Normal and Friction Forces:
} At discontinuous boundary points where a tubular domain meets a notched domain (see Fig. \ref{fig:modeling2}), concentrated normal  $\boldsymbol{F}_{N}(s)$ and concentrated friction forces $\boldsymbol{F}_{F}(s)$   must be taken into account and calculated. 
% At the points of discontinuity, the concentrated forces (i.e., concentrated normal force $\boldsymbol{F}_{N}(s)$ and concentrated friction force $\boldsymbol{F}_{F}(s)$) 
As mentioned, these forces depend on the angle $\phi_t$-- defined in (\ref{eq: change in angle}) indicating the change  in the tangent direction of the tendon $\boldsymbol{t}_{\text{tendon}}$ when it enters/exits to/from the  tubular domain-- and can
% as follows: 
% are a function of the change in angle $\phi_t$ in the tangent direction of the tendon curve. It can
also be described using the tendon kinematics tuple function  as below:
\begin{align}
\boldsymbol{F}_N(s)=
\begin{cases}
    \underbrace{\boldsymbol{n}_{\text{tendon}}(\mathcal{L}_i) \sin{\phi_t(\mathcal{L}_i)}}_{\boldsymbol{\mathcal{P}_N(\mathcal{L}_i)}} T(\mathcal{L}_i^{-}) \hspace{24
pt} s \in \mathfrak{L}\\
     \underbrace{\boldsymbol{n}_{\text{tendon}}(\mathcal{R}_i) \sin{\phi_t(\mathcal{R}_i)}}_{\boldsymbol{\mathcal{P}_N(\mathcal{R}_i)}} T(\mathcal{R}_i^{-}) \hspace{20
pt} s \in \mathfrak{R}
\end{cases}
\label{eq:concentrated normal force}
\end{align}
% It is also possible to obtain concentrated friction force $\boldsymbol{F}_{F}(s)$:
\begin{align}
\boldsymbol{F}_F(s)=
\begin{cases}
    \underbrace{-\boldsymbol{t}_{\text{tendon}}(\mathcal{L}_i) \hl{\mu}\cos{\phi_t(\mathcal{L}_i)}}_{\boldsymbol{\mathcal{P}_F(\mathcal{L}_i)}} T(\mathcal{L}_i^{-}) \hspace{24pt} s \in \mathfrak{L}\\
     \underbrace{-\boldsymbol{t}_{\text{tendon}}(\mathcal{R}_i)\hl{\mu} \cos{\phi_t(\mathcal{R}_i)}}_{\boldsymbol{\mathcal{P}_F(\mathcal{R}_i)}} T(\mathcal{R}_i^{-}) \hspace{20pt} s \in \mathfrak{R}
\end{cases}
\label{eq:concentrated frictional force}
\end{align}
where, the terms $\boldsymbol{\mathcal{P}_N}$ and $\boldsymbol{\mathcal{P}_F}$ are called the ``\textit{projection factors of concentrated forces}" that project the tendon tension $\boldsymbol{T}(s)$ at discontinuous boundary points onto the concentrated normal   $\boldsymbol{F}_N(s)$ and the the  friction  $\boldsymbol{F}_F(s)$ forces, respectively.
Detailed derivation of (\ref{eq:concentrated normal force}) and (\ref{eq:concentrated frictional force}) can be found in Appendix \ref{derivation2}.
% In (\ref{eq:distributed normal force}) and (\ref{eq:concentrated normal force}), $\boldsymbol{\mathcal{P}_n}$ and $\boldsymbol{\mathcal{P}_N}$ are defined as the ``projection factor" that can ``project" tendon tension to distributed normal force and concentrated normal force, respectively. Similarly, in (\ref{eq:distributed friction force}) and (\ref{eq:concentrated frictional force}),  $\boldsymbol{\mathcal{P}_f}$ and $\boldsymbol{\mathcal{P}_F}$ are defined as the ``projection factor" that can ``project" tendon tension to distributed friction force and concentrated frictional force, respectively.
% Therefore, we can use a general physically-intuitive form to express the distributed forces and concentrated forces on the tendon as follow:
% The equations (\ref{eq:distributed normal force}) and (\ref{eq:concentrated normal force}) introduce the terms $\boldsymbol{\mathcal{P}_n}$ and $\boldsymbol{\mathcal{P}_N}$ as the "projection factors" that facilitate the "projection" of tendon tension onto distributed normal force and concentrated normal force, respectively. Similarly, equations (\ref{eq:distributed friction force}) and (\ref{eq:concentrated frictional force}) define $\boldsymbol{\mathcal{P}_f}$ and $\boldsymbol{\mathcal{P}_F}$ as the "projection factors" responsible for projecting tendon tension onto distributed friction force and concentrated frictional force, respectively.

\textbf{Tendon Force Tuple Function:}
Investigating (\ref{eq:tendon tension})--(\ref{eq:concentrated frictional force}) and utilizing the defined projection factors  $\boldsymbol{\mathcal{P}_f}$ and $\boldsymbol{\mathcal{P}_F}$,  $\boldsymbol{\mathcal{P}_n}$ and $\boldsymbol{\mathcal{P}_N}$, we are able to adopt a physically-intuitive formulation to express either of the distributed  and concentrated forces acting on the tendon  using the following compact form:
\begin{align} 
\label{eq:general force}
\boldsymbol{force}
=\boldsymbol{\mathcal{P}}\boldsymbol{T}({\boldsymbol{\mathcal{G}(s)}})
\end{align}
where $\boldsymbol{\mathcal{P}}$ represents either of $\boldsymbol{\mathcal{P}_n}$, $\boldsymbol{\mathcal{P}_N}$, $\boldsymbol{\mathcal{P}_f}$, and $\boldsymbol{\mathcal{P}_F}$ and  the variable  $\boldsymbol{force}$ can be either of $\boldsymbol{F}_{F}$, $\boldsymbol{f}_f$,  $\boldsymbol{F}_{N}$ and $\boldsymbol{f}_n$.

\begin{remark}
\label{remark4} 
\textit{Equation (\ref{eq:general force}) provides a generalized and intuitive formulation that enhances our understanding of the distributed forces and concentrated forces acting on the tendon within a TD-CM featuring discontinuous cross-section geometry. The physical intuition behind this  formulation  can simply be  explained as follows: as tendon tension $\boldsymbol{T}$ is transmitted from the base of a TD-CM to a specific point along the tendon, it first undergoes an exponential   and concentrated tension loss-- calculated by {$\boldsymbol{\mathcal{G}}$} term. This tension is then ``projected" onto the normal and tangent directions of the tendon curve and can be calculate using the $\boldsymbol{\mathcal{P}}$ term. Of note, this ``projection" is captured by the tendon kinematics tuple function $(\boldsymbol{p}_{t}, 
\boldsymbol{t}_{\text{tendon}}, \boldsymbol{n}_{\text{tendon}},  \kappa_t, \phi_t )$.
% , which characterizes the tendon's position, tangent vector, normal vector, curvature, and angle change in the tangent vector at the given point. 
% By incorporating these factors, we can gain a physically-intuitive understanding of how forces are distributed and concentrated along the tendon in a TD-CMs with FPs that experiences tension loss. 
Therefore, besides the tendon's kinematics tuple function, in this work, we also introduce a force tuple function $( \boldsymbol{T}, \boldsymbol{f}_n, \boldsymbol{f}_f, \boldsymbol{F}_{N}, \boldsymbol{F}_{F})$. This tuple function is a collection of five functions that can comprehensively and intuitively  represent and describe the tendon's statics and process of tension loss in a TD-CM with FPs.}
\end{remark}

% \begin{remark}
% \label{remark4} 
% \textit{Equation (\ref{eq:general force}) generalizes and provides a physically-intuitive understanding of distributed forces and concentrated forces on the tendon for TD-CMs with discontinuous cross-section geometries or flexure patterns. When tendon tension is transmitted from the base of TD-CMs to a point on the tendon, it first experiences an exponential tension loss and concentrated tension loss and then  is ``projected" to the normal and tangent directions of the tendon curve. The ``projection factor" is characterized by the tendon kinematics tuple function $(\boldsymbol{p}_{t}, 
% \boldsymbol{t}_{\text{tendon}}, \boldsymbol{n}_{\text{tendon}},  \kappa_t, \phi_t )$.}
% \end{remark}

\subsection{Tendon's Moment Tuple Function:}
% The moment that a tendon applies to the midline of the TD-CM, which is distributed throughout the structure, can be attributed to three main sources. First, the anchor point force, as shown in Fig. \ref{fig:modeling1}, contributes to this moment. Second, the internal distributed forces also play a role. Lastly, the concentrated forces due to discontinuity contrbuite to the moment as well.
The total moment acting on the midline  of the TD-CM % that is exerted by the forces acting on a tendon (represented by the force tuple function)
 % is distributed across the manipulator's structure,
 can be attributed to three primary sources including   the anchor point force acting on the manipulator's distal end as well as    the internal distributed and the concentrated forces acting on the tendon  (see Fig. \ref{fig:modeling1}). The following sections describe how to calculate either of these moment components and represent it by a moment tuple function.
 % this  forces within the TD-CM also contribute to the moment. Lastly, the concentrated forces resulting from discontinuities in the structure also contribute to the overall moment.
\subsubsection{Anchor Point Moment}
The moment created by the anchor point force $\boldsymbol{m}_{\text{point}}$ is determined by taking the cross product of the arm vector and the anchor point force vector $\boldsymbol{T}(L)$:
\begin{gather}
\boldsymbol{m}_{\text{point}}(s)
=\nonumber\underbrace{(\boldsymbol{p}_{t}(L)-\boldsymbol{p}(s))}_{arm}\times\underbrace{\boldsymbol{T}(L)}_{anchor \hspace{3pt}point\hspace{3pt}force}\\
\nonumber
= (\boldsymbol{p}(s)-\boldsymbol{p}_{t}(L)) \times \boldsymbol{t}_\text{tendon}(L) T(\mathcal{R}_N^{+}) \\  e^{-\mu  (\theta_t(L)-\theta_t(\mathcal{R}_N))}
\label{eq:moment equations1}
\end{gather}

% The total moment $\boldsymbol{m}_{total}$ exerted by the tendon on the TD-CM's midline is composed of the contribution of the anchor point force, distributed normal forces along the tendon channel, and concentrated point forces. The moment $\boldsymbol{m}_{a}$ generated by the anchor point force is calculated as the cross product of the arm vector and the anchor point force vector:

% To express the moment $\boldsymbol{m}_{n}$ resulting from the distributed normal forces and distributed friction forces, one can integrate the cross product of the arm vector and the distributed normal forces, which can be represented as:

\subsubsection{Distributed Moment}
To calculate the moment that arises from the distributed normal  and friction forces $\boldsymbol{m}_{\text{distributed}}$, one can integrate the cross product of the arm vector and the distributed normal $\boldsymbol{f}_n$ and  friction  $\boldsymbol{f}_f$ forces. This can be expressed as:
\begin{gather}
\boldsymbol{m}_{\text{distributed}}(s)
=\nonumber\int_s^{L}\underbrace{(\boldsymbol{p}_{t}(\xi)-\boldsymbol{p}(s))}_{arm}\times\underbrace{(\boldsymbol{f}_n(\xi)+\boldsymbol{f}_f(\xi))}_{distributed\hspace{3pt}forces} d\xi \\
=\nonumber\int_s^{L}(\boldsymbol{p}_{t}(\xi)-\boldsymbol{p}(s)) \times (\boldsymbol{n}_{\text{tendon}}(\xi)-\mu  \boldsymbol{t}_{\text{tendon}}(\xi))\\
 \kappa_t(\xi) T(\mathcal{R}_i^{+})  e^{-\mu  (\theta_t(\xi)-\theta_t(\mathcal{R}_i))} d\xi
\label{eq:moment equations2}
\end{gather}

\subsubsection{Concentrated Point Moment}
The moment produced by the concentrated point forces $\boldsymbol{m}_{\text{concentrated}}$ is obtained by taking the cross product of the arm vector and the concentrated point force vectors (i.e., $\boldsymbol{F}_{N}, \boldsymbol{F}_{F}$):
\begin{gather}
\boldsymbol{m}_{\text{concentrated}}(s)
=\nonumber\sum_{\xi\in \mathfrak{L}\cup \mathfrak{R}}\underbrace{(\boldsymbol{p}_{t}(\xi)-\boldsymbol{p}(s))}_{arm}\times\underbrace{(\boldsymbol{F}_{N}(\xi)+\boldsymbol{F}_{F}(\xi))}_{concentrated \hspace{3pt}force}\\
=\nonumber\sum_{\xi\in \mathfrak{L}}[(\boldsymbol{p}_{t}(\xi)-\boldsymbol{p}(s))\times 
 (\boldsymbol{n}_{\text{tendon}}(\mathcal{\xi}) \sin{\phi_t(\mathcal{\xi})}\\-\boldsymbol{t}_{\text{tendon}}(\mathcal{\xi}) \cos{\phi_t(\mathcal{\xi})})
 \nonumber T(\mathcal{\xi}^{-})]\\
 \nonumber+\sum_{\xi\in \mathfrak{R}}[(\boldsymbol{p}_{t}(\xi)-\boldsymbol{p}(s))\times 
 (\boldsymbol{n}_{\text{tendon}}(\mathcal{\xi}) \sin{\phi_t(\mathcal{\xi})}\\-\boldsymbol{t}_{\text{tendon}}(\mathcal{\xi}) \cos{\phi_t(\mathcal{\xi})}) T(\mathcal{\xi}^{-})]
\label{eq:moment equations3}
\end{gather}
\subsubsection{Total Moment}
The total moment on the TD-CM's midline $\boldsymbol{m}_{\text{total}}$  can be expressed as:
\begin{align}\label{eq:total moment}
    \boldsymbol{m}_{\text{total}}(s)=\boldsymbol{m}_{\text{point}}(s)+\boldsymbol{m}_{\text{distributed}}(s)+\boldsymbol{m}_{\text{concentrated}}(s)
\end{align}

\begin{remark}
\label{remark5} 
\textit{Besides the tendon's kinematics and force tuple functions, we introduce a moment tuple function denoted as $(\boldsymbol{m}_{\text{point}} , \boldsymbol{m}_{\text{distributed}}, \boldsymbol{m}_{\text{concentrated}}, \boldsymbol{m}_{\text{total}})$,
which is a collection of four functions that represent the effects of tendon loads on the TD-CM's midline total moment.}
\end{remark}

\subsection{Equilibrium Equation and Iterative Solution Algorithm}
To analyze the deformation and shape of a TD-CM experiencing tension-loss under various tendon tensions, it is necessary to formulate an equilibrium equation that balances the forces and moments exerted by the tendons and the constituting material of the TD-CM. To this end, the total moment $\boldsymbol{m}_{\text{total}}(s)$ calculated in (\ref{eq:total moment}) can be mapped to the deformation behavior of the TD-CM using the following static equilibrium equation:
\begin{align} \label{eq:equilbrium}
\boldsymbol{m}_{\text{total}}(s)&=\boldsymbol{K}_{w}(s)\omega_b(s) 
% \nonumber 
% \\
% \boldsymbol{K}_{w}(s)&= EI(s) 
\end{align}
where $\boldsymbol{K}_{w}(s)= EI(s)$ is the stiffness matrix, $E$ is the Young's modulus, and $I(s)$ is the piecewise second moment of area function defined in (\ref{eq:2nd moment of area}). 
Notably,  changing the geometry of FPs--  using the parametric form of superellipse function in (\ref{eq:superellipse equation2})-- directly affects on the  stiffness matrix and deformation behavior of manipulator. Therefore, the proposed modeling framework can readily be used to analyze and design various types of discontinuous TD-CMs.

\subsection{Holistic Description of the Modeling Framework and the Iterative Solution Algorithm}

In the previous sections, we defined the following four tuple functions  on their respective domains:
(1) The tendon kinematics tuple function $(\boldsymbol{p}_{t}, 
\boldsymbol{t}_{\text{tendon}}, \boldsymbol{n}_{\text{tendon}},  \kappa_t, \phi_t )$  defined on the domain $\boldsymbol{X}$;
(2) The forces tuple function $( \boldsymbol{T}, \boldsymbol{f}_n, \boldsymbol{f}_f, \boldsymbol{F}_{N}, \boldsymbol{F}_{F})$ defined on the domain $\boldsymbol{\Xi}$;  
(3) The moment tuple function $(\boldsymbol{m}_{\text{point}} , \boldsymbol{m}_{\text{distributed}}, \boldsymbol{m}_{\text{concentrated}}, \boldsymbol{m}_{\text{total}})$  defined on the domain $\boldsymbol{\Upsilon}$; and (4)
the TD-CM's kinematics tuple function  $(\mathcal{R}, \boldsymbol{p})$  defined on the domain $\boldsymbol{\Omega}$.
The relationships between these four tuple functions and their involved  parameters can also be described using the following four functional mappings $\boldsymbol{\mathcal{F}}, \boldsymbol{\mathcal{M}},\boldsymbol{\mathcal{K}}, \boldsymbol{\mathcal{Q}}$ that can intuitively be used for modeling and solving their nonlinear interrelated relationships. 

First, \textit{Function $\boldsymbol{\mathcal{F}}$ maps \textbf{tendon kinematics} domain $\boldsymbol{X}$ to the \textbf{tendon statics}} {co}-domain $\boldsymbol{\Xi}$:
\begin{gather}
    \nonumber\boldsymbol{\mathcal{F}}: \boldsymbol{X} \longrightarrow \boldsymbol{\Xi},\\
    (\boldsymbol{p}_{t}, 
\boldsymbol{t}_{\text{tendon}}, \boldsymbol{n}_{\text{tendon}},  \kappa_t, \phi_t )\mapsto(\boldsymbol{T}, \boldsymbol{f}_n, \boldsymbol{f}_f, \boldsymbol{F}_{N}, \boldsymbol{F}_{F})
\label{eq:tendon kinematics and tendon statics}
\end{gather}
 Tendon kinematics in a TD-CM involves understanding and quantifying the movement of the tendon as they transmit forces to the flexible structure of the manipulator. This includes determining the position, orientation, and shape of the tendon at different points along the TD-CM's length asdescribed in (\ref{eq:psotion vector of tendon})-(\ref{eq: change in angle}). Tendon statics involves the examination and analysis of the forces that are applied to the tendons responsible for actuating the deformation of a TD-CM {(see (\ref{eq:tendon tension})-(\ref{eq:concentrated frictional force}))}. 
Tendon kinematics and tendon statics of TD-CM modeling can be connected and defined using ``tension loss factor" (i.e., $\boldsymbol{\mathcal{G}}$) and ``projection factors" (i.e., $\boldsymbol{\mathcal{P}_n}$, $\boldsymbol{\mathcal{P}_f}$, $\boldsymbol{\mathcal{P}_N}$, $\boldsymbol{\mathcal{P}_F}$) to obtain a physically-intuitive mathematical formulation as shown in(\ref{eq:general force}).

Second, \textit{Function $\boldsymbol{\mathcal{M}}$ maps the \textbf{tendon statics} domain  $\boldsymbol{\Xi}$ to  the \textbf{TD-CM statics}} {co}-domain $\boldsymbol{\Upsilon}$:
\begin{gather}
    \nonumber\boldsymbol{\mathcal{M}}: \boldsymbol{\Xi} \longrightarrow \boldsymbol{\Upsilon},\\
(\boldsymbol{T}, \boldsymbol{f}_n, \boldsymbol{f}_f, \boldsymbol{F}_{N}, \boldsymbol{F}_{F})\mapsto(\boldsymbol{m}_{\text{point}}, \boldsymbol{m}_{\text{distributed}}, \boldsymbol{m}_{\text{concentrated}}, \boldsymbol{m}_{\text{total}})
\label{eq: tendon statics and TD-CM statics}
\end{gather}
TD-CM statics refers to the analysis of the static equilibrium conditions and forces acting on a TD-CM. {\textit{Function $\boldsymbol{\mathcal{M}}$ indicates  that}} how the moments exerted by a tendon on the midline of the TD-CM (described by (\ref{eq:moment equations1})-(\ref{eq:total moment})) can be attributed
to tendon statics equations summarized in (\ref{eq:tendon tension})-(\ref{eq:concentrated frictional force}).

Third, \textit{Function $\boldsymbol{\mathcal{K}}$ maps the \textbf{TD-CM statics} domain $\boldsymbol{\Upsilon}$ to \textbf{TD-CM kinematics}} {co}-domain $\Omega$:  
\begin{gather}
    \nonumber\boldsymbol{\mathcal{K}}: \boldsymbol{\Upsilon} \longrightarrow \boldsymbol{\Omega},\\
    (\boldsymbol{m}_{\text{point}}, \boldsymbol{m}_{\text{distributed}}, \boldsymbol{m}_{\text{concentrated}}, \boldsymbol{m}_{\text{total}})\mapsto(\mathcal{R}, \boldsymbol{p})
\label{eq: TD-CM statics and TD-CM kinematics}
\end{gather}
TD-CM statics {refers to the moments acting on the midline of the TD-CM (i.e., (\ref{eq:tendon tension})-(\ref{eq:concentrated frictional force}))} while  TD-CM kinematics {refers to} the midline curvature, bending angle, and the robot's shape/deformation behavior using (\ref{eq:rigid-body transformation matrix}).
 % to \yl{\sout{strain} angular or linear rate of changes of the TD-CM's configuration with respect to $s$} (i.e., TD-CM kinematics (see)). 
 TD-CM kinematics and TD-CM statics of continuum robot modeling can be connected by material mechanics (e.g., Cosserat rod theory) and (\ref{eq:equilbrium}). 

Finally, \textit{Function $\boldsymbol{\mathcal{Q}}$ maps \textbf{TD-CM kinematics} domain $\boldsymbol{\Omega}$ to the \textbf{tendon kinematics}} {co}-domain $\boldsymbol{X}$: 
\begin{gather}
    \nonumber\boldsymbol{\mathcal{Q}}: \boldsymbol{\Omega} \longrightarrow \boldsymbol{X},\\
    (\mathcal{R}, \boldsymbol{p})\mapsto(\boldsymbol{p}_{t}, 
\boldsymbol{t}_{\text{tendon}}, \boldsymbol{n}_{\text{tendon}},  \kappa_t, \phi_t )
\label{eq: TD-CM kinematics and tendon kinematics}
\end{gather}
The relationship between TD-CM kinematics {(i.e., (\ref{eq:rigid-body transformation matrix}))} and tendon kinematics {(described by (\ref{eq:psotion vector of tendon})-(\ref{eq: change in angle})) lies in how the motion of the tendons affects the overall kinematics of the TD-CM. The tendon actuates and controls the deformation of TD-CM, which, in turn, determines the overall motion and configuration of the TD-CM.
By understanding the kinematics of the tendon, such as its length changes and orientations, it is possible to determine how these kinematic parameters influence the motion and deformation of the TD-CM as a whole. This information is essential for designing and controlling the TD-CM, as it allows us to determine the required tendon configurations and actuation patterns to achieve desired TD-CM kinematics.

% \subsection{Iterative Solution Algorithm}
To solve the proposed nonlinear modeling framework  expressed by the aforementioned four functions (\ref{eq:tendon kinematics and tendon statics})-(\ref{eq: TD-CM kinematics and tendon kinematics}), we propose an iterative solution approach outlined in Algorithm \ref{alg:algorithm1}. As summarized, in  the first step, algorithm first calculates tendon kinematics (\ref{eq:psotion vector of tendon})-(\ref{eq: change in angle}) in a straight andunbent configuration using \textit{Function $\boldsymbol{\mathcal{Q}}$ expressed in (\ref{eq: TD-CM kinematics and tendon kinematics})}}. The obtained tendon kinematics defined in (\ref{eq:psotion vector of tendon})-(\ref{eq: change in angle}) will then be used to initialize the iteration process and solve the  proposed model described based on the  four mapping functions. Particularly, mappings $\boldsymbol{\mathcal{F}}, \boldsymbol{\mathcal{M}},\boldsymbol{\mathcal{K}}, \boldsymbol{\mathcal{Q}}$  are iteratively solved   in order until the following  convergence conditions (i.e., stopping criteria) are satisfied:
\begin{align}
\label{eq:convergence}
\begin{cases}
\| \boldsymbol{p}^{k+1}(L)-\boldsymbol{p}^{k}(L)\|_2 > \epsilon_{tip,th} \\
\{\sum_{i=1}^m\| \boldsymbol{p}^{k+1}(s_i)-\boldsymbol{p}^{k}(s_i)\|_2\}/{m} > \epsilon_{shape,th}
\end{cases}
\end{align}

% \yl{where $\boldsymbol{p}^{k}(L)$  represents the tip position of the TD-CM  at the $k^{th}$ iteration and the left side of the first equation  represents the   tip position increment between two successive iterations $k+1$ and $k$.  Also, $\boldsymbol{p}^{k}(s_i)$ denotes the $i^{th}$ cross-section position at the $k^{th}$ iteration and the left side of the second equation is the average increment of total N calculated cross-section positions between  two successive iterations, which is a measurement of shape difference. $\epsilon_{tip,th}$ and $\epsilon_{shape,th}$ are predefined error thresholds.}

where $\boldsymbol{p}^{k}(L)$ denotes the tip position of the TD-CM at the $k^{th}$ iteration, and the left side of the first equation represents the increment in tip position between two consecutive iterations $k+1$ and $k$.
Similarly, $\boldsymbol{p}^{k}(s_i)$ represents the $i^{th}$ cross-section position at the $k^{th}$ iteration, and the left side of the second equation signifies the average increment in the total {$m$}-calculated shape of the manipulator through cross-section positions between two consecutive iterations. 
Moreover, $\epsilon_{tip,th}$ and $\epsilon_{shape,th}$ are predefined error thresholds that will be used as the stopping criteria.

 \begin{algorithm}[t]
\caption{Solving the  Proposed  Mappings }
\label{alg:algorithm1}
\SetAlgoLined
\nonl\textbf{Input:}  $T_{\text{base}}$, $n$, $\mu$

\nonl\textbf{Initialization:}

$(\boldsymbol{p}_{t}^0, 
\boldsymbol{t}_{\text{tendon}}^0, \boldsymbol{n}_{\text{tendon}}^0,  \kappa_t^0, \phi_t^0 )$ $\longleftarrow$ Initialize TD-CM kinematics [straight TD-CM shape];

$k \longleftarrow 0$;
% $\theta^0 \longleftarrow \pi/2$;

\nonl\textbf{Iterative Process:}

\Do{{$\| \boldsymbol{p}^{k+1}(L)-\boldsymbol{p}^{k}(L)\|_2 > \epsilon_{tip,th}$} or
$\{\sum_{i=1}^N\| \boldsymbol{p}^{k+1}(s_i)-\boldsymbol{p}^{k}(s_i)\|_2\}/{N} > \epsilon_{shape,th}$}{ [$\boldsymbol{T}^{k+1}$, $\boldsymbol{f}_n^{k+1}$, $\boldsymbol{f}_f^{k+1}$, $\boldsymbol{F}_{N}^{k+1}$, $\boldsymbol{F}_{F}^{k+1}$] $\longleftarrow$ Calculate function mapping $\boldsymbol{\mathcal{F}}$ (\ref{eq:tendon kinematics and tendon statics}) [$\boldsymbol{p}_{t}^{k}$, 
$\boldsymbol{t}_{\text{tendon}}^{k}$, $\boldsymbol{n}_{\text{tendon}}^{k}$,  $\kappa_t^{k}$, $\phi_t^{k}$];

[$\boldsymbol{m}_{\text{point}}^{k+1}$, $\boldsymbol{m}_{\text{distributed}}^{k+1}$, $\boldsymbol{m}_{\text{concentrated}}^{k+1}$, $\boldsymbol{m}_{\text{total}}^{k+1}$] $\longleftarrow$ Calculate function mapping $\boldsymbol{\mathcal{M}}$ (\ref{eq: tendon statics and TD-CM statics}) [$\boldsymbol{T}^{k+1}$, $\boldsymbol{f}_n^{k+1}$, $\boldsymbol{f}_f^{k+1}$, $\boldsymbol{F}_{N}^{k+1}$, $\boldsymbol{F}_{F}^{k+1}$];

[$\mathcal{R}^{k+1}$, $\boldsymbol{p}^{k+1}$] $\longleftarrow$ Calculate function mapping $\boldsymbol{\mathcal{K}}$ (\ref{eq: TD-CM statics and TD-CM kinematics}) [$\boldsymbol{m}_{\text{point}}^{k+1}$, $\boldsymbol{m}_{\text{distributed}}^{k+1}$, $\boldsymbol{m}_{\text{concentrated}}^{k+1}$, $\boldsymbol{m}_{\text{total}}^{k+1}$];

[$\boldsymbol{p}_{t}^{k+1}$, 
$\boldsymbol{t}_{\text{tendon}}^{k+1}$, $\boldsymbol{n}_{\text{tendon}}^{k+1}$,  $\kappa_t^{k+1}$, $\phi_t^{k+1}$] $\longleftarrow$ Calculate function mapping $\boldsymbol{\mathcal{Q}}$ (\ref{eq: TD-CM kinematics and tendon kinematics}) [$\mathcal{R}^{k+1}$, $\boldsymbol{p}^{k+1}$];

$k \longleftarrow k + 1$;
}

\nonl\hl{\textbf{Output:}} $\boldsymbol{p}$, $ \mathcal{R}$

\end{algorithm}

\section{Simulation and Experimental Validations}
\subsection{Simulation Studies}
To evaluate the performance of the proposed generic
 framework, we conducted the following two types of  simulations in MATLAB (MathWorks, Inc.) on TD-CMs with geometries described in Section \ref{sec: superellipse modeling} and shown in Fig. \ref{fig:modeling1}. The considered TD-CMs of the two simulations share the following identical design parameters: $L=53$ mm, $l_{base}=5$ mm, $l_{tip}=4$ mm, $l_n=2$ mm, $l_t=4$ mm, $t_d=1.6$ mm, $N=8$, $R=3.92$ mm, $r=1.55$ mm.  Also, we used Young's modulus of $E$ = 1600 MPa and the following parameters as the stopping criteria in Algorithm \ref{alg:algorithm1}: $\epsilon_{tip,th}$ = 0.10 mm and $\epsilon_{shape,th}$ = 0.05 mm with $m$ = 1060. The geometric design of FPs were determined based on the exponent value of the SEF  in (\ref{eq:superellipse equation1}).
Of note, 
the proposed modeling framework is generic and the used parameters can be changed depending on the application.

\subsubsection{Analysis of SEF-base FP geometry on TD-CM deformation Behavior}
The first type of simulations was performed to investigate the effects of three different  geometries for FPs defined based on the SEF   (denoted as \textit{$S1^{\star}\sim S3^{\star}$}  corresponding  with exponent value $n$ = 1, 3, 5 in (\ref{eq:superellipse equation1}), respectively) on the deformation behavior of a TD-CM, the midline curvature distribution, tendon-tension transmission loss, as well as distributed  and concentrated forces.
We also analyzed the  relationships among input tendon tension, the selected exponent values, and TD-CM's tip displacement. 
\begin{figure*}[!t] 
\centering
\includegraphics[width=0.98\linewidth]{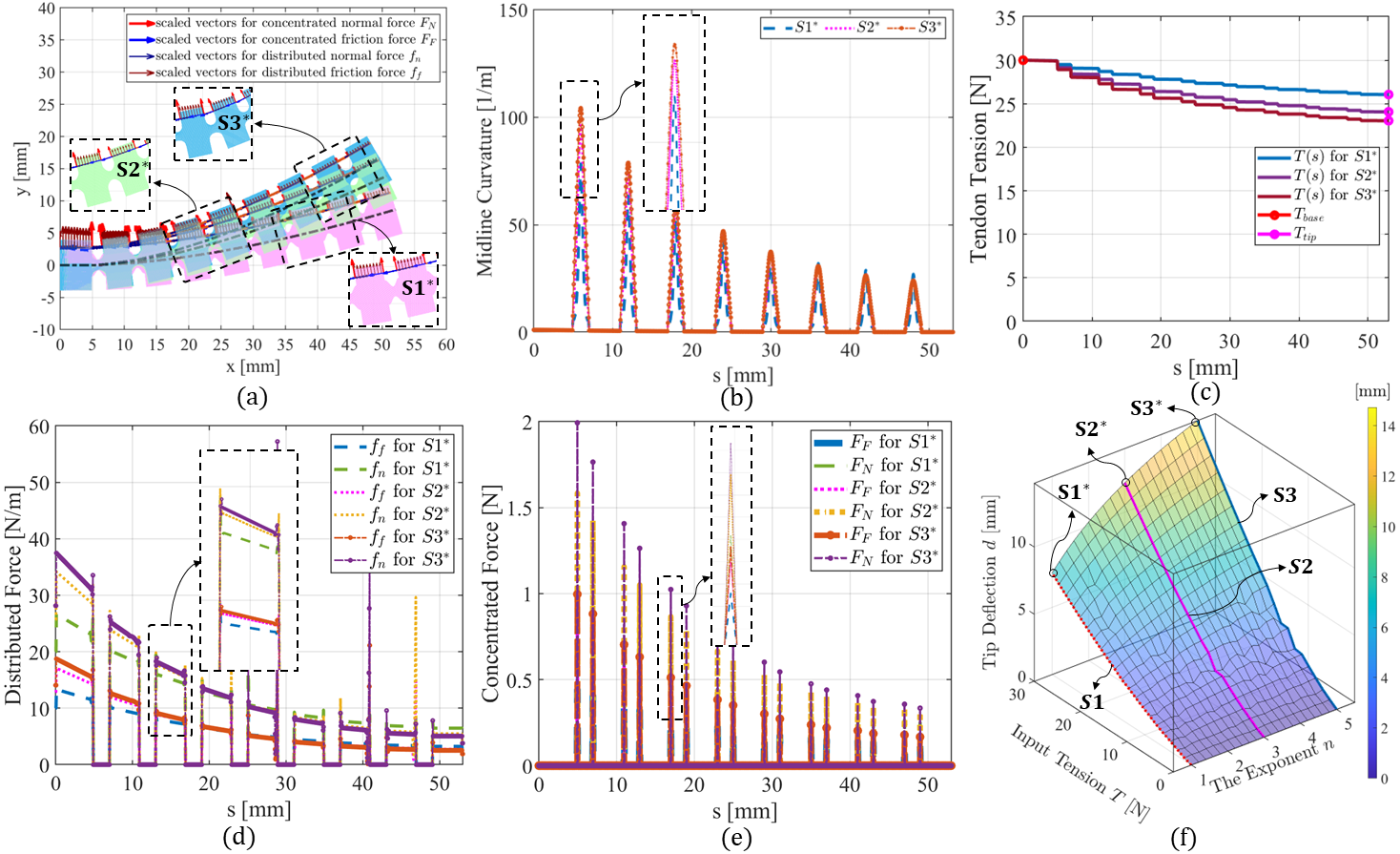}
\centering
\caption{Simulation results of TD-CMs with three different superellipse-geometric FPS (i.e., \textit{$S1^{\star}\sim S3^{\star}$}) actuated with an identical input tension $T_{\text{base}}=$ 30 N and considered friction coefficient $\mu$ = 0.5: (a) Comparison and illustration of the TD-CMs' deformation behaviors; (b) Comparison of the midline curvatures along   midline arc length $s$ for the considered TD-CMs; (c) Comparison of the tendon tension distribution between the scenarios \textit{$S1^{\star}\sim S3^{\star}$};
(d) Comparison of internal distributed forces along the tendon channel for the three selected TD-CMs;
(e) Comparison of concentrated forces at the discontinuity between \textit{$S1^{\star}\sim S3^{\star}$};
(f) The surface represents the \textit{design space}  showing the space of all possible exponent value $n$ and tip displacement $d$ when the input tension is changing from  $T=$ 0 $\sim$ 30 N. The design space also includes \textit{design curves \textit{$S1\sim S3$}}  representing the tip Deflection $d$ of  TD-CMs with  specific FP geometries corresponding to exponent $n$=1, 3, and 5 in (\ref{eq:superellipse equation1}). }
\label{fig:sim1}
\end{figure*}
\begin{figure*}[!t] 
\centering
\includegraphics[width=0.98\linewidth]{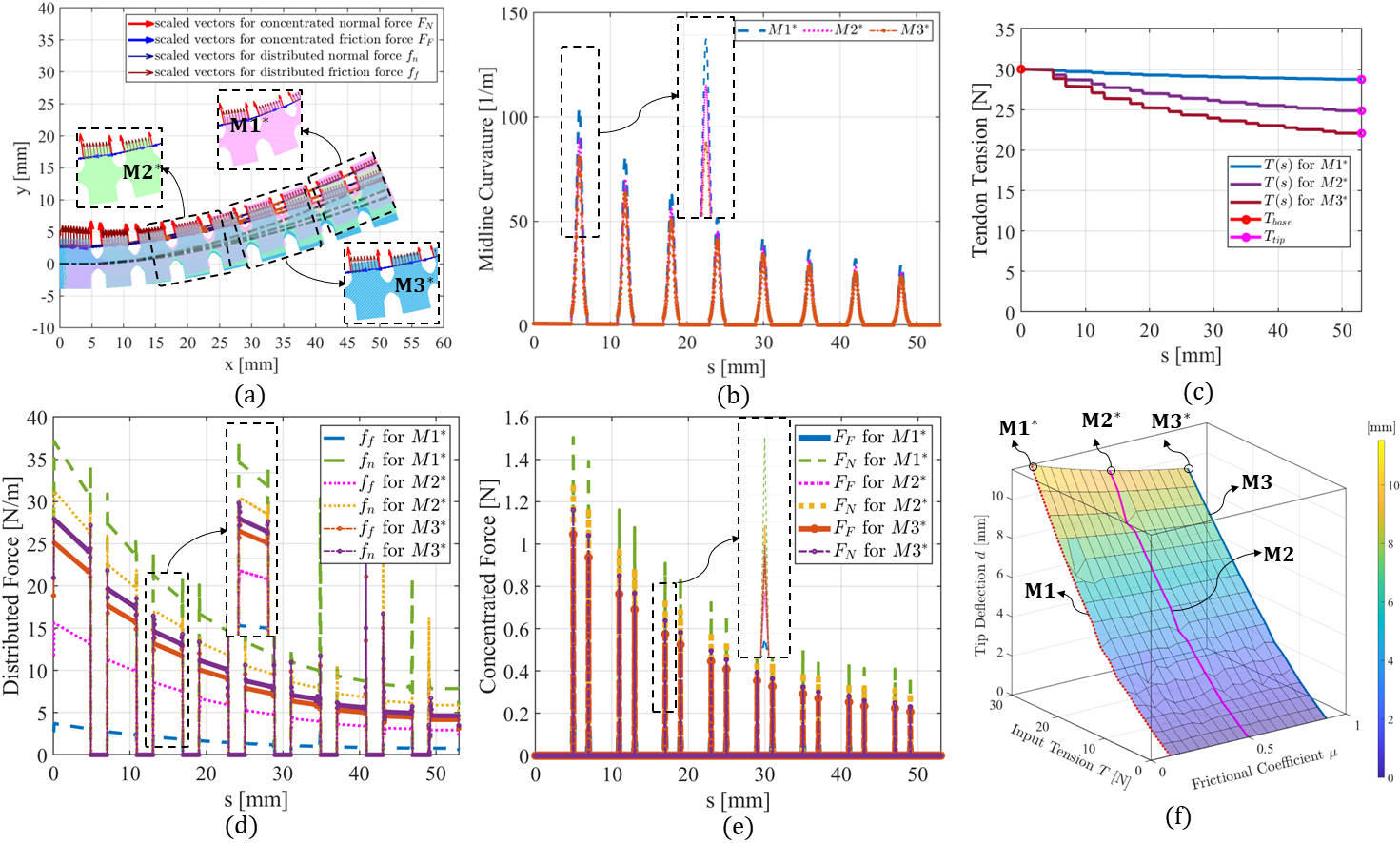}
\centering
\caption{Simulation results of TD-CMs with three different friction coefficients (i.e., \textit{$M1^{\star}\sim M3^{\star}$}) actuated with an identical input tension $T_{\text{base}}=$ 30 N: (a) Comparison and illustration of the TD-CMs' deformation behaviors; (b) Comparison of the midline curvatures along midline arc length 
 $s$ for the considered friction coefficients; (c) Comparison of the tendon tension distribution between these scenarios \textit{$M1^{\star}\sim M3^{\star}$};
(d) Comparison of internal distributed forces along the tendon channel for the three selected friction coefficients;
(e) Comparison of concentrated forces at the discontinuity between \textit{$M1^{\star}\sim M3^{\star}$};
(f) The surface represents the \textit{design space}  showing the space of all  possible friction coefficients $\mu$, and tip displacement $d$ when the input tension is changing from  $T_{\text{base}}=$ 0 $\sim$ 30 N. The design space also includes \textit{design curves} \textit{$M1\sim M3$} representing the tip Deflection $d$ of  TD-CMs with  specific friction coefficients corresponding to $\mu$= 0.1, 0.5, and 1  in (\ref{eq:tendon tension}).}
\label{fig:sim2}
\end{figure*}
Figure \ref{fig:sim1}(a) compares the deformation behaviors of the three different scenarios \textit{$S1^{\star}\sim S3^{\star}$} calculated using the proposed modeling framework and iterative solution algorithm under an arbitrarily selected input tendon tension (i.e., $T_{\text{base}}=$ 30 N) and  friction coefficient $\mu$ = 0.5. Figure \ref{fig:sim1}(b) illustrates the curvature distribution along  TD-CMs midline for the considered scenarios. 
% Fig. \ref{fig:sim1}(c) shows the tendon curvature distribution and wrap angle associated with \textit{$S1\sim S3$}. 
Figure \ref{fig:sim1}(c) compares the tendon tension distribution between the considered scenarios \textit{$S1^{\star}\sim S3^{\star}$}.
Figure \ref{fig:sim1}(d) and Fig. \ref{fig:sim1}(e) show the internal distributed forces along the tendon channel and concentrated forces at the discontinuity points of each FP (shown in Fig. \ref{fig:modeling2}), respectively. Figure \ref{fig:sim1}(f) also illustrates the ``\textit{design space}" together with a comparison of the ``\textit{design curves}" associated with \textit{$S1^{\star}\sim S3^{\star}$}. 
 Generated by the proposed model and iterative solution algorithm, the \textit{design space}  demonstrates the space of all possible exponent value $n$ and tip displacement $d$ when the input tension is changing from  $T=$ 0 $\sim$ 30 N.   As indicated within the \textit{design space}, a \textit{design curve}  is a subset of this space representing the deformation behavior (i.e., Tip Deflection $d$) of a TD-CM with a specific FP geometry (defined by exponent $n$ of SEF function)  under different input tensions. In other words, the aforementioned \textit{design space} of $n$, $d$, and $T$, shown in Fig. \ref{fig:sim1}(f), will be reduced to a \textit{design curve} due to the specific geometric pattern imposed on TD-CM's deformation. This figure also shows the considered  special scenarios of \textit{$S1^{\star}, S2^{\star}$} and \textit{$S3^{\star}$} located on three demonstrated \textit{design curves} (i.e., \textit{$S1, S2$} and \textit{$S3$}).
 The \textit{design space} and \textit{design curve}, therefore, demonstrate how  for a TD-CM with discontinuous cross-section geometries, the deformation behavior, flexibility or steerability can directly be affected by the SEF exponent value $n$ in (\ref{eq:superellipse equation1}).

\subsubsection{Analysis of Friction on Tension Loss and Deformation Behavior of TD-CMs}
The second type of simulations investigates the impact of three different friction coefficients (i.e., scenarios \textit{$M1^{\star}\sim M3^{\star}$} with $\mu$ = 0.1, 0.5, 0.9 in (\ref{eq:tendon tension}), respectively) on the deformation behavior of a TD-CM, the midline curvature distribution, tendon-tension transmission loss, distributed forces, concentrated forces, 
and the relationships among input tendon tension, friction coefficient, and TD-CM's tip displacement. For our simulations, without loss of generality, we considered a TD-CM with an FP geometry with $n$ = 2 in (\ref{eq:superellipse equation1}). 
Figure \ref{fig:sim2} shows simulation results of this TD-CM  actuated with an identical input tension of $T=$ 30 N while experiencing three different friction coefficients (i.e., \textit{$M1^{\star}\sim M3^{\star}$}).
Figure \ref{fig:sim2}(a) provides a comparison and illustration of the TD-CMs' deformation behaviors calculated using the proposed modeling framework and iterative solution algorithm under the three selected scenarios \textit{$M1^{\star}\sim M3^{\star}$}. 
Figure \ref{fig:sim2}(b) compares the midline curvatures along $s$ for the considered friction coefficients.
Also, Fig. \ref{fig:sim2}(c) shows a comparison of the tendon tension distribution among these scenarios \textit{$M1^{\star}\sim M3^{\star}$}.
Figure \ref{fig:sim2}(d) illustrates internal distributed forces along the tendon channel for the three selected friction coefficients.
Figure \ref{fig:sim2}(e) compares concentrated forces at the discontinuity points of each FP (shown in Fig. \ref{fig:modeling2}) among \textit{$M1^{\star}\sim M3^{\star}$}. Moreover, Fig. \ref{fig:sim2}(f) illustrates the ``\textit{design space}" representing the relationship among all possible friction coefficients $\mu$ ranging from 0.1 to 0.9 and tip displacement $d$ when the input tension is changing from  $T=$ 0 $\sim$ 30 N. \textit{$M1\sim M3$} ``\textit{design curves"} also represent the aforementioned relationship generated by specifying the friction coefficient $\mu$ in (\ref{eq:tendon tension}). Similar to Fig. \ref{fig:sim1}(f), here, the \textit{design curve}  is also a subset of the \textit{design space} representing the deformation behavior  of one specific TD-CM  under different input tensions and friction coefficients. In other words, the aforementioned \textit{design space} of $\mu$, $d$, and $T$, shown in Fig. \ref{fig:sim2}(f), will be reduced to a \textit{design curve} due to the specific friction coefficient imposed on TD-CM's deformation. This figure also shows the considered  special scenarios of \textit{$M1^{\star}, M2^{\star}$} and \textit{$M3^{\star}$} located on three demonstrated \textit{design curves} (i.e., \textit{$M1, M2$} and \textit{$M3$}).
\subsection{Experimental Studies}
\begin{figure*}[!t] 
\centering
\includegraphics[width=0.9\linewidth]{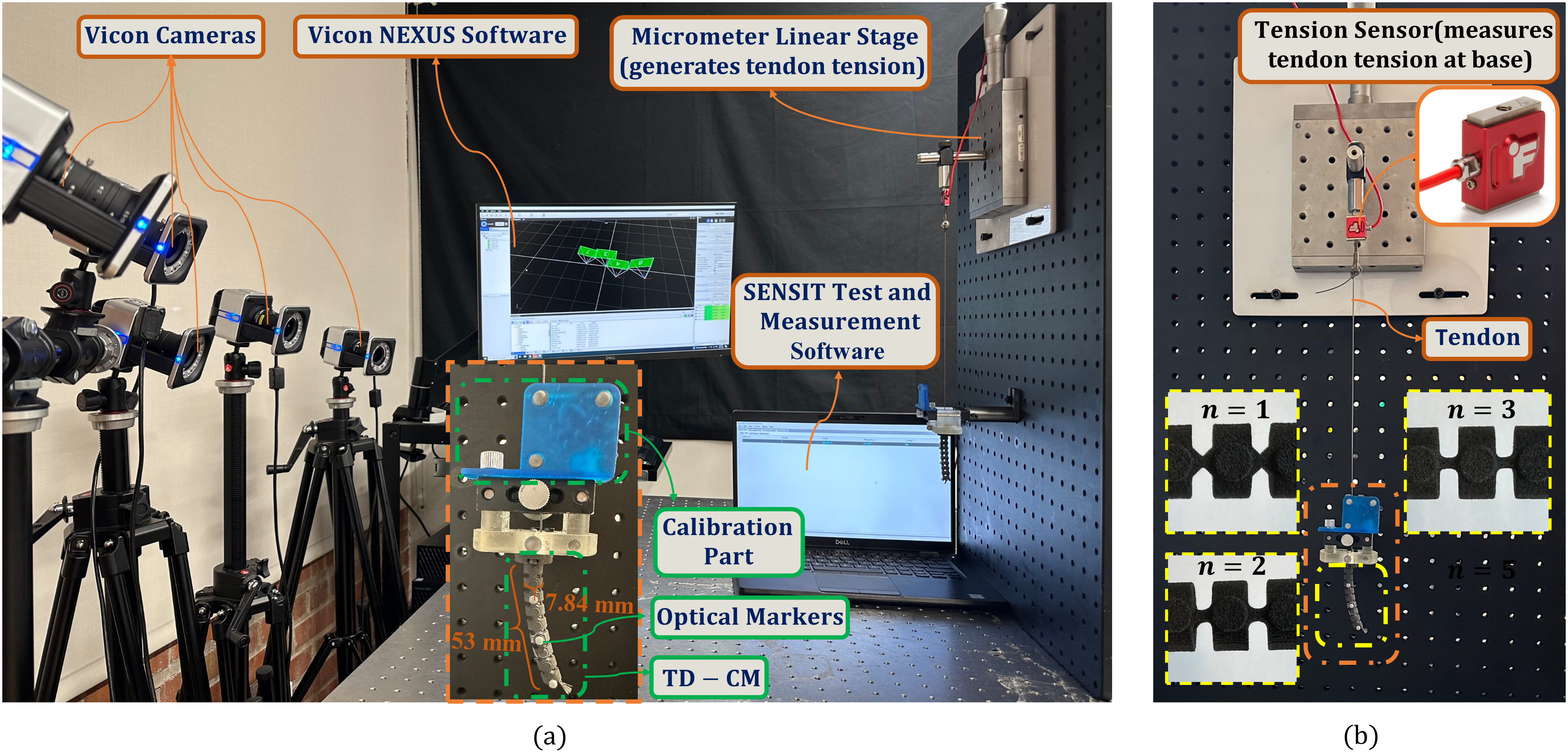}
\centering
\caption{(a) Overview of the utilized experimental
setup  comprising of an additively manufactured TD-CM, tension sensor for measuring the tendon tension,  linear stage for actuation of the manipulator,  and the Vicon motion capture system used to measure the deformation behavior of the manipulator during actuation; (b) Figure shows the geometry of the considered FPs for the fabricated TD-CMs corresponding to the exponent value $n=$ 1, 2, and 3 in the SEF.}
\label{fig:exp_setup}
\end{figure*}
To experimentally assess the efficacy of the proposed framework in  analyzing the deformation behavior of a TD-CM with various FP geometries (determined by the exponent value $n$ of the SEF) experiencing tension loss, we designed a setup  depicted in Fig. \ref{fig:exp_setup}. Particularly, for these experiments, we additively manufactured three TD-CMs  using elastic and durable Nylon 12 Powder (Formlabs,  Young's modulus $E$ = 1600 MPa) through selective laser sintering (SLS) 3D printing approach. The TD-CM utilized in the experiments feature  discontinuous cross-section geometries defined by the SEF (illustrated in Fig. \ref{fig:modeling1} and Fig. \ref{fig:exp_setup}). The geometric parameters of the TD-CM, presented in Fig. \ref{fig:modeling1}, are as follows: $L=53$ mm, $l_{base}=5$ mm, $l_{tip}=4$ mm, $l_n=2$ mm, $l_t=4$ mm, $t_d=1.4$ mm, $N=8$, $R=3.92$ mm, $r=1.55$ mm.
Actuation of the TD-CMs was achieved using a low-stretch coated wire rope (34235T26, McMaster-Car) with a diameter of 0.03 inch. The friction coefficient was measured through the experimental method outlined in \cite{yang2020}, yielding a friction coefficient of $\mu=0.17$.
% The wire rope is highly flexible and threaded through the tendon channel, with a knot made at the tip of the TD-CM. 
% The path of the tendon is defined by the parameter in (\ref{eq:psotion vector of tendon}), specifically $\boldsymbol{r}_{t}(s)=$ 2.75 mm.
To provide a specific tendon tension as input, a load cell (LSB201, FUTEK Advanced Sensor Technology, Inc.) was connected to the opposite end of the wire rope, which was attached to a linear stage mechanism (M-UMR12.40, Newport Corporation). Also,  we used the following parameters as the stopping criteria in Algorithm \ref{alg:algorithm1}: $\epsilon_{tip,th}$ = 0.10 mm and $\epsilon_{shape,th}$ = 0.05 mm with $m$ = 1060.

% This experimental setup enabled the evaluation of the proposed framework's effectiveness in simulating and analyzing the deformation behavior of the TD-CM with different flexural patterns, providing valuable insights into its performance.

% The setup consists of the proposed TD-CM and a Vicon motion capture system, 

% To capture the motion/deformation behavior of the TD-CM, Vicon motion capture system with five cameras was set around the experimental setup and a calibration part with three optical markers was utilized to define the coordinate frame visually in the Vicon NEXUS software. After calibration, the Vicon system can capture the motions of optical markers attached to the TD-CM's body. In this study, the three optical markers were attached to the locations of 15 mm, 33 mm and 53 mm from its base cross section to a cross section on which the markers were attached.
% In order to record the deformation characteristics of the proposed TD-CM, Vicon motion capture system comprising of six cameras was arranged around the experimental setup. A calibration part (see Fig. \ref{fig:exp_setup}(b)) consisting of three optical markers was employed to visually establish the coordinate frame in the Vicon NEXUS software (shown in Fig. \ref{fig:exp_setup})(a). Following calibration, the Vicon system could capture the movements of the optical markers affixed to the body of the TD-CM. In this investigation, three optical markers were placed at locations 15 mm, 33 mm, and 53 mm from the base cross section of the TD-CM to a cross section on which the markers were attached.
As shown in Fig. \ref{fig:exp_setup},  Vicon motion capture system (Vero, Vicon Motion Systems, Ltd.), comprising five cameras, was positioned around the experimental setup to capture the motion and deformation behavior of the TD-CM.  Particularly, the Vicon system utilized a calibration part with three optical markers to establish the visual coordinate frame within the Vicon NEXUS software (Vicon Motion Systems, Ltd.). After calibration, the Vicon system accurately tracked the movements of the optical markers attached to the TD-CM's body. Specifically, three optical markers were affixed at positions 15 mm, 33 mm, and 53 mm from the base cross section to a designated cross section.
% To record the deformation characteristics of the proposed TD-CM, a Vicon motion capture system consisting of six cameras was arranged around the experimental setup. A calibration part (depicted in Fig. \ref{fig:exp_setup}(b)) containing three optical markers was used to establish the coordinate frame visually in the Vicon NEXUS software (illustrated in Fig. \ref{fig:exp_setup}(a)). Following calibration, the Vicon system accurately captured the movements of the optical markers attached to the TD-CM's body. For this investigation, three optical markers were strategically positioned at distances of 15 mm, 33 mm, and 53 mm from the base cross section of the TD-CM to the corresponding cross section where the markers were affixed.
This setup enabled the comprehensive recording of the TD-CM's motion and deformation used for analyzing its behavior and validating the proposed modeling framework. 

\begin{figure*}[!t] 
\centering
\includegraphics[width=1\linewidth]{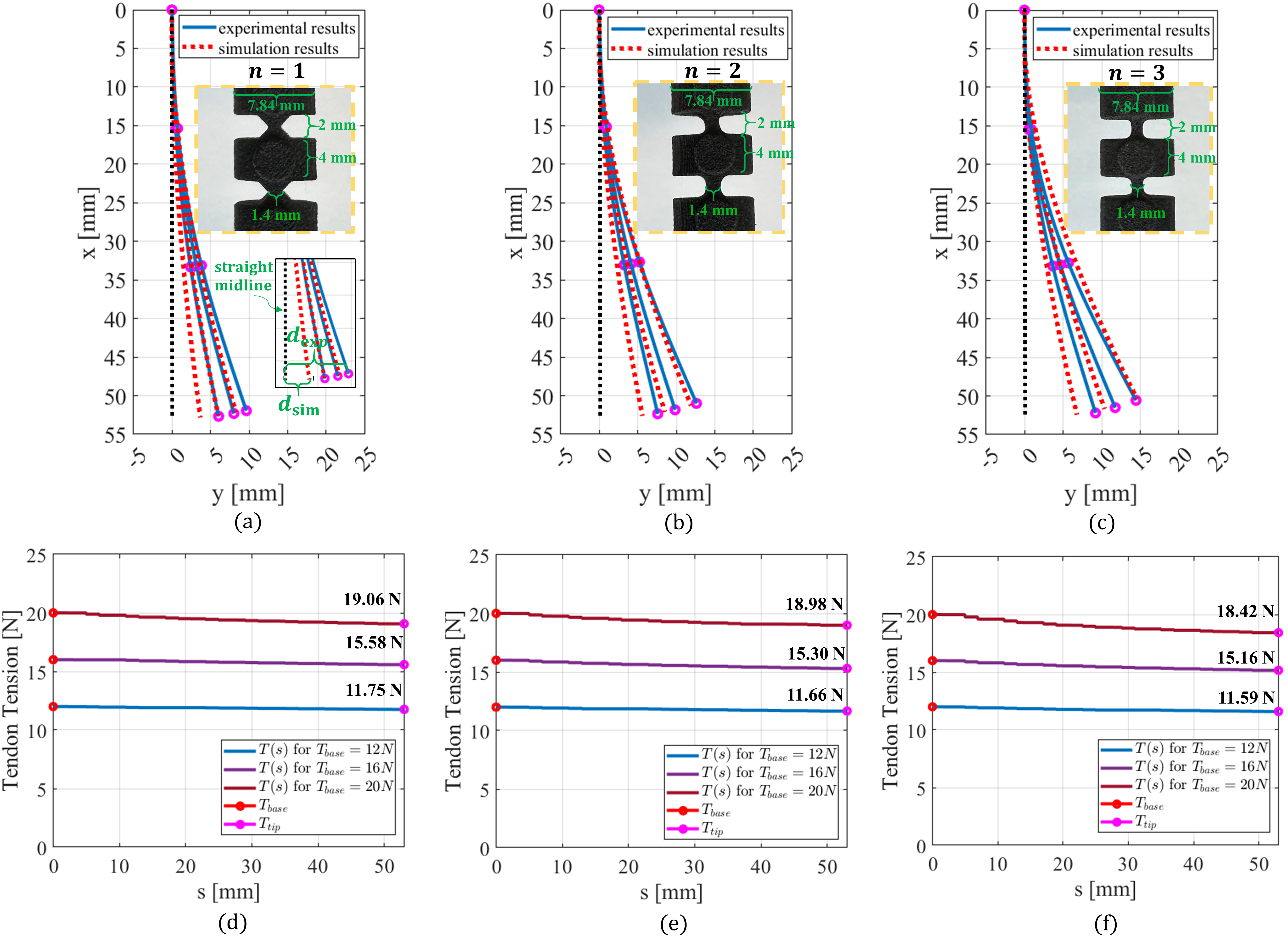}
\centering
\caption{Figures (a)-(c) compare the experimental results and the model predictions of the midline shape for TD-CMs with three different FPs based on the SEF geometry defined by exponent values $n=$ 1, 2, and 3  at three different input tensions $T_{\text{base}}=$ 12, 16, and 20 N, respectively. Figures  \ref{fig:exp1}(d)-(f) also show the corresponding estimated tendon tension loss in the considered cases calculated by the proposed modeling framework.}
% Comparison of the proposed TD-CM's midline shape between the experimental results and model prediction for four selected different superellipse-geometric flexure patterns (i.e., exponent value $n=$ 1, 2, 3) with input tendon tension $T_{\text{base}}=$ 12 $\sim$ 20 N with a 4 N increment: (a) Experimental results for TD-CM notch design with exponent value $n=$ 1; (b) Experimental results for TD-CM notch design with exponent value $n=$ 2; (c) Experimental results for TD-CM notch design with exponent value $n=$ 3; (d) Tendon tension distribution for TD-CM notch design with exponent value $n=$ 1; (e) Tendon tension distribution for TD-CM notch design with exponent value $n=$ 2; (f) Tendon tension distribution for TD-CM notch design with exponent value $n=$ 3.}
\label{fig:exp1}
\end{figure*}

% Fig. \ref{fig:exp1} shows a comparison between the experimental results and model prediction of the midline shape of TD-CMs with four different superellipse-geometric flexure patterns (i.e., exponent value $n=$ 1, 2, 3, 5). Since the captured motion data of the TD-CM shape is discrete locations of the optical markers, cubic spline data interpolation technique (see \textit{cscvn} function in MATLAB) was utilized to interpolate these discrete points to create a smooth curve to represent the TD-CM's midline shape.
% Of note, the magenta circles indicate the locations of optical markers attached to the TD-CM's body, while the red dashed curve corresponds to the aforementioned cubic spline interpolation.
% As shown in TABLE I, the mean absolute percentage error (MAPE), root mean square error (RMSE), and mean absolute error (MAE) are used to measure the accuracy of the tip deflection of the TD-CM in three repeated trials between experimental results and model predictions. As shown in Fig. \ref{fig:exp2}, \textit{Shape error mean} is used to represent the deviation of the entire TD-CM's shape in three repeated trials between experimental results and model predictions. \textit{Tip deflection mean} is the average value of the tip deflection in three repeated experimental trials.
In Fig. \ref{fig:exp1}(a)-(c), a comparison is presented between the experimental results and the model predictions of the midline shape for TD-CMs with three different FPs based on the SEF geometry defined by exponent values $n=$ 1, 2, and 3  at three different input tensions $T_{\text{base}}=$ 12, 16, and 20 N. Of note, we limited our measurements at these values as tensions more than 20 N could break the additively manufactured TD-CMs and tensions less than 12 N did not create a significant deformation behavior for the manipulators.  Since the captured motion data consists of discrete locations of the optical markers attached to the TD-CM's body, a cubic spline data interpolation technique was employed  using the \textit{cscvn} function in MATLAB (Mathworks Inc.) to create a smooth curve representing the TD-CM's midline shape.
The magenta circles in the figure indicate the positions of the optical markers, while the red dashed curve corresponds to the aforementioned cubic spline interpolation. Figures  \ref{fig:exp1}(d)-(f) also show the estimated tendon tension loss in the considered cases calculated by the proposed modeling framework.
To compare the obtained experimental results with the proposed model predictions, we used TD-CM's tip deflection $d$  by calculating the distance between the tip and midline axis of the actuated TD-CM with respect to  its unactuated straight configuration. We then defined the mean absolute percentage error (MAPE) (\ref{eq:MAPE}) and mean absolute error (MAE) (\ref{eq:MAE}) and presented them in TABLE \ref{table:1}:
\begin{align}
\label{eq:MAPE}
\textit{MAPE} = \frac{1}{n_{\text{trial}}}\sum_{i=1}^{n_{\text{trial}}} | \frac{d_{\text{i,sim}}-d_{\text{i,exp}}}{d_{\text{i,exp}}}| 
\end{align}
\begin{align}
\label{eq:MAE}
\textit{MAE} = \frac{1}{n_{\text{trial}}}\sum_{i=1}^{n_{\text{trial}}} |d_{\text{sim}}-d_{\text{i,exp}}| 
\end{align}

where $d_{\text{sim}}$ is the calculated TD-CM's tip deflection using the proposed modeling framework and $d_{\text{i,exp}}$ is the TD-CM's tip deflection of $i^{th}$ trial in experiment. These metrics were computed based on three repeated trials (i.e, $n_{\text{trial}}$ = 3) for the  tip deflection with $T_{\text{base}}=$ 12 N, 16 N and 20 N. 

% \yl{We use tip deflection measurement (i.e., the distance between TD-CM's tip and TD-CM's midline axis in its straight configuration) to provide  quantitative measures of the agreement between the experimental results and the model predictions.}

% To evaluate the accuracy of the tip deflection of the TD-CM, the mean absolute percentage error (MAPE) and mean absolute error (MAE) were calculated and presented in TABLE \ref{table:1}. These metrics were computed based on three repeated trials for the \yl{\sout{maximum}} tip deflection with $T_{\text{base}}=$ \yl{12 N, 16 N and} 20 N

Moreover, we defined $E_{\text{shape}}$ representing the deviation of the entire TD-CM's shape between model prediction and experiments as follows:
% \yl{\sout{-- as well as the \textit{Tip deflection mean}}}-- \yl{which is the average value of discrete points along the TD-CM's midline}-- for $T_{\text{base}}=$ 20 N. 
\begin{align}
\label{eq:shape error}
E_{\text{shape}} = \{\sum_{i=1}^m\| \boldsymbol{p}_{\text{sim}}(s_i)-\boldsymbol{p}_{\text{exp}}(s_i)\|_2\}/{m} 
\end{align}
where $\boldsymbol{p}_{\text{sim}}(s_i)$ represents the $i^{th}$ TD-CM's cross-section position obtained from the model, and $\boldsymbol{p}_{\text{exp}}(s_i)$ indicates the $i^{th}$ TD-CM's cross-section position from experiments.
This equation signifies the average difference between the considered  total $m$ = 1060  points along the manipulator' geometry calculated through the proposed model and obtained by experiments.

Fig. \ref{fig:exp2}  compares $MAE$ and $E_{\text{shape}}$ with the averaged measured tip deflection at the performed experiments ($d_{\text{mean, exp}}$).
Of note, these metrics were computed based on three repeated experimental trials and compared to the model predictions. 
Overall, the comparison between experimental results and model predictions provides insights into the accuracy and performance of the proposed framework in capturing the shape and tip deflection behavior of the TD-CMs with different FPs.

\section{Discussion}

As opposed to the previous literature designing a TD-CM with discrete geometry using presumed (e.g., {\cite{Handed2021RAL}})  or time-consuming FEA-obtained (e.g., \cite{goury2018fast}, \cite{adagolodjo2021coupling}) FPs, the shown results in Fig. \ref{fig:sim1}, Fig. \ref{fig:sim2}, and Fig.\ref{fig:exp1} clearly demonstrate that the proposed generic SEF-based approach can readily be used for the design and  thorough analysis of the geometrical patterns on the deformation behavior and tension loss of TD-CMs. 

\begin{table}
\centering
\begin{threeparttable}
\caption{Experimental results for the TD-CM's tip deflection at $T_{\text{base}}=$ 12 N, 16 N and 20 N in three repeated trials with four different FPs (i.e., exponent value $n=$ 1, 2, 3). $\sigma$ represents the standard deviation of TD-CM's tip deflection between experiments and model outputs.} 
\label{table:1}
\setlength\tabcolsep{0pt} % make LaTeX figure out intercolumn spacing
\begin{tabular*}{0.95\columnwidth}{@{\extracolsep{\fill}} c | c c c c}
% \begin{tabular*}
\toprule
Input Tension & Exponent Value $n$ & 1 & 2 & 3 \\
\midrule
\multirow{6}{4em}{$T_{\text{base}}=$ 12 N} 
& $d_{\text{sim}}$ & 3.68 mm & 5.61 mm & 6.79 mm \\
& $d_{\text{mean, exp}}$ & 6.34 mm & 7.85 mm & 9.25 mm \\
& $\sigma$ & 0.13 mm & 0.58 mm & 0.40 mm \\
& MAPE & 41.81\% & 28.38\% & 26.53\% \\
& MAE & 2.65 mm & 2.23 mm & 2.45 mm \\
& $E_{\text{shape}}$ & 0.88 mm & 0.82 mm & 0.75 mm \\
\midrule
\multirow{6}{4em}{$T_{\text{base}}=$ 16 N} 
& $d_{\text{sim}}$ & 5.99 mm & 8.51 mm & 10.37 mm \\
& $d_{\text{mean, exp}}$ & 8.14 mm & 10.36 mm & 12.08 mm \\
& $\sigma$ & 0.31 mm & 0.26 mm & 0.19 mm \\
& MAPE & 26.43\% & 17.70\% & 14.04\% \\
& MAE & 2.15 mm & 1.84 mm & 1.70 mm \\
& $E_{\text{shape}}$ & 0.67 mm & 0.71 mm & 0.61 mm \\
\midrule
\multirow{6}{4em}{$T_{\text{base}}=$ 20 N} 
& $d_{\text{sim}}$ & 8.29 mm & 11.91 mm & 14.58 mm \\
& $d_{\text{mean, exp}}$ & 9.99 mm & 12.84 mm & 14.88 mm \\
& $\sigma$ & 0.36 mm & 0.23 mm & 0.60 mm \\
& MAPE & 17.00\% & 7.30\% & 2.62\% \\
& MAE & 1.71 mm & 0.94 mm & 0.40 mm \\
& $E_{\text{shape}}$ & 0.55 mm & 0.60 mm & 0.84 mm \\
\bottomrule
\end{tabular*}
\smallskip
\scriptsize
\end{threeparttable}
\end{table}	
From Fig. \ref{fig:sim1}(a) and Fig. \ref{fig:sim1}(b) and despite using identical input tendon tension of 30 N, friction coefficient, and constituting material for the three considered TD-CMs with various superellipse-geometric FPs (i.e., scenarios $S1^{\star}$ to $S3^{\star}$), distinct deformation behaviors and midline curvature distributions are observed. This clearly highlights the influence of different exponent values $n$ on the  overall shape of a notched TD-CM.In other words, by adjusting the value of the exponent $n$ in an SEF, the flexibility and workspace of the TD-CM can be customized and tailored to specific application requirements. 
For instance, in Fig. \ref{fig:sim1}(a), when the exponent value $n$ is increased from 1 to 5, the TD-CM's maximum deformation increases by approximately 50$\%$ from 7.5 mm (i.e., $S1^{\star}$ scenario) to 15 mm (i.e., $S3^{\star}$ scenario). When $n=1$, the curve resembles a rhombus with corners, while for $n=2$, the curve becomes a regular ellipse. When $n=5$, the curve appears similar to a rectangle with rounded corners. Therefore, as the exponent value increases, the notches become more rectangular in shape, resulting in more materials being removed from the manipulator that directly affect    the second moment of area $I(s)$ about the bending axis of the TD-CM  in (\ref{eq:2nd moment of area}) and subsequently stiffness matrix $\boldsymbol{K}_{w}$ in (\ref{eq:equilbrium}). This explains why TD-CMs with identical material properties and different FPs featuring larger exponent values ($n$) exhibit greater flexibility and larger deformation of about 15 mm compared to 7.5 mm for smaller exponent values.
Furthermore, as can be observed in Fig. \ref{fig:sim1}(a) and Fig. \ref{fig:sim1}(b), the curvature distribution of the tendon in $S3^{\star}$ are larger than those in $S1^{\star}$ and $S2^{\star}$. This explains why $S3^{\star}$ experiences higher cumulative distributed friction and tension loss (about 23$\%$ loss or 7 N), as shown in Fig. \ref{fig:sim1}(c)-(e).

% As clearly observed in Fig. \ref{fig:sim1}(d), when a tendon is fully constrained within a TD-CM tubular section, it becomes curved during the
% manipulator’s deformation, which explains why it experiences an exponential tension loss within this section. In contrast,  the tendon located in the notched section without any constraints remains a straight line which does not create any distributed normal forces or friction. It is worth noting that concentrated forces occur at discontinuity points as shown in Fig. \ref{fig:sim1}(e), where the tubular section
% meets the notched section since there is a sudden change in the tangent direction of the tendon curve. As shown in Fig. \ref{fig:sim1}(f), although three selected geometries $S1^{\star}\sim S3^{\star}$ are distinct, their corresponding ``\textit{design curves}" all lie on the  ``design space" and are a subset of the design space. 
% In Fig. \ref{fig:sim1}(d), it is evident that when a tendon is fully constrained within the tubular section of the TD-CM, it becomes curved during the manipulator's deformation. 
In Fig. \ref{fig:sim1}(d), it is clear that as the manipulator undergoes a deformation during actuation, the parts of tendon that are fully constrained within     the tubular sections of the TD-CM experience a curved shape (see Fig. \ref{fig:modeling2}).
These piece-wise curved shapes for the constrained parts of tendon indeed explain the exponential tension loss experienced by the tendon within the tubular section. On the other hand, the unconstrained parts of tendon located inside the notched sections of TD-CM would obtain a straight/linear shape, resulting in zero distributed normal forces or friction within these domains.
As observed in Fig. \ref{fig:sim1}(e), it is important to note that concentrated forces occur at the points of discontinuity (Fig. \ref{fig:modeling2}), where a tubular section meets the notched section. This abrupt change in the tangent direction of the tendon curve leads to the occurrence of concentrated forces.
Furthermore, Fig. \ref{fig:sim1}(f) demonstrates that although the three selected geometries ($S1^{\star}$ to $S3^{\star}$) exhibit distinct geometries for FPs, their corresponding ``design curves" all lie within the ``design space" and are a subset of it. This clearly features   the use and importance  of  the proposed modeling framework and the ``design space" concept for optimal and intuitive design of a TD-CM for various applications. In other words,  any curve located between the extreme scenarios of $S1^{\star}$ and $S3^{\star}$ on this space represents a potential FP geometry and its effect on the deformation behavior of the manipulator in the presence of various input tensions.   

 \begin{figure}[!t] 
\centering
\includegraphics[width=\linewidth]{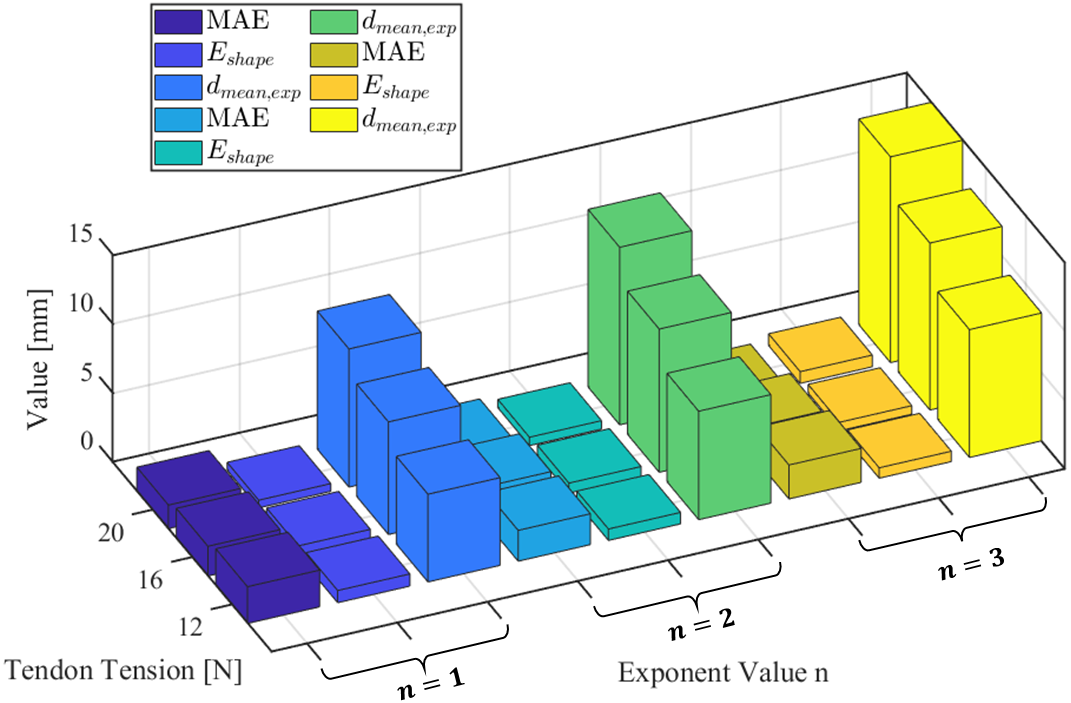}
\centering
\caption{ Figure compares $MAE$ and $E_{\text{shape}}$ with the averaged measured tip deflection in three repeated trials ($d_{\text{mean, exp}}$) for $T_{\text{base}}=$ 12 N, 16 N and 20 N  and three different FPs (i.e., exponent value $n=$ 1, 2, 3).}
% Statistical analysis for experimental  for the maximum TD-CM's tip deflection at $T_{\text{base}}=$ 20 N and shape error with four different superellipse-geometric flexure patterns (i.e., exponent value $n=$ 1, 2, 3, 5) in three repeated trials.
\label{fig:exp2}
\end{figure}

From Fig. \ref{fig:sim2}, it can be observed that TD-CMs with identical flexural patterns (presented by  $S2^{\star}$ scenario  in Fig. \ref{fig:sim1}), material constitution, and input tension ($T_{\text{base}}$ = 30 N) exhibit different deformation behaviors, midline curvature distributions, tension loss, internal distributed forces, and concentrated forces under different friction coefficient conditions ($M1^{\star}$ to $M3^{\star}$, corresponding to $\mu$ = 0.1, $\mu$ = 0.5, and $\mu$ = 0.9).
Similar to the effects of  FPs on deformation behavior of TD-CMs demonstrated in Fig. \ref{fig:sim1}, the influence of friction coefficients on the deformation behavior is also clear (see Fig. \ref{fig:sim2}(a) and Fig. \ref{fig:sim2}(b)). As can be observed,  when the friction coefficient $\mu$ is increased from 0.1 to 0.9, the TD-CM's maximum deformation decreases by about 18$\%$ from 11.5 mm (i.e., $M3^{\star}$ scenario) to 9.5 mm (i.e., $M3^{\star}$ scenario).
Compared with the the effect of FPs on the tension loss of the  TD-CM with $S2^{\star}$ geometry shown in Fig.  \ref{fig:sim1}(c) (i.e., about 23$\%$ or 7 N in tension loss), investigation of Fig. \ref{fig:sim2}(c) shows that the change in the friction coefficient has a greater impact on the transmission loss of tendon tension in $M3^{\star}$ scenario (i.e., about 27$\%$ or 8 N  tension loss).
Furthermore, in Fig. \ref{fig:sim2}(d) and Fig. \ref{fig:sim2}(e), higher TD-CM deformations (associated with $M1^{\star}$) result in higher distributed   and concentrated normal forces but lower distributed and concentrated friction forces due to the smaller friction coefficient ($\mu$ = 0.1). This behavior differs from the one shown in Fig. \ref{fig:sim1}(d) and Fig. \ref{fig:sim1}(e), where higher normal forces are associated with larger friction forces. Similar to Fig. \ref{fig:sim1}(f), the ``design curves" corresponding to the three selected friction coefficients ($M1^{\star}$ to $M3^{\star}$) in Fig. \ref{fig:sim2}(f) lie within the ``design space" and are a subset of it.
These results demonstrate the varying influence of friction coefficients on the behavior of TD-CMs, emphasizing the importance of considering both geometry  and friction conditions during the optimal design of a TD-CM.

% As shown in Fig. \ref{fig:exp1}, Fig. \ref{fig:exp2} and TABLE \ref{table:1}, despite different superellipse-geometric flexure patterns of the proposed TD-CM, the proposed model reliably and accurately obtained the robot's deformation behavior in three repeated trials. The small prediction errors, including MAPE $<$ 8\%, MAE/RMSE $<$ 1 mm, and \textit{shape error mean} $<$ 1.6 mm, validated the accuracy of the proposed model and solution algorithm for modeling TD-CMs with discontinuous cross-section geometries for exponent value $n=$ 2, 3 and 5. It is worth noting that for the case of exponent value $n=$ 1, the TD-CM has a very small deformation in which its maximum tip deflection is below 10 mm (shown in Fig. \ref{fig:exp2}), which explains why it has a relatively large prediction error (i.e., MAPE $<=$ 17\%).
% The design and manufacture of TD-CMs can be costly, and the intricate nature of the robot can pose challenges for scaling up production. However, by utilizing the proposed accurate model for designing control algorithms and predicting robot performance with varying design parameters, the design and fabrication of TD-CMs and their actuation systems can be optimized before the fabrication process. This can potentially help to reduce costs and improve efficiency in the fabrication of TD-CM.

Fig. \ref{fig:exp1}, Fig. \ref{fig:exp2}, and TABLE \ref{table:1} demonstrate that the proposed generic model is able to  capture the deformation behavior of the considered  TD-CMs in three repeated trials. Overall, as can be observed from these figures and table,  the considered errors (i.e., MAPE, MAE, $E_{\text{shape}}$)  increase when a smaller  input tension  is used for actuation and/or TD-CM is built of a  smaller   SEF exponent value. Of note, in these cases,    TD-CM experiences a smaller deformation (i.e., $<\sim$ 10 mm) causing a relatively larger calculated error between the proposed model and measured results. On the other hand, when the robot experiences a deformation  $>\sim$ 10 mm, these errors would reduce to less than 17\%.  For example, as summarized in TABLE \ref{table:1}, the prediction errors for $T_{\text{base}}=$ 20 N are very small, with overall MAPE $<$ 8 \%, MAE $<$ 1 mm, and $E_{\text{shape}}$ $<$ 1.6 mm, for  TD-CMs with discontinuous cross-section geometries and exponent values of $n=$ 2 and 3.   Also, for the case of $T_{\text{base}}=$ 16 N and exponent values of $n=$ 3, MAPE $<$ 14\%, MAE $<$ 1.7 mm, and $E_{\text{shape}}$ $<$ 0.61 mm were calculated for maximum calculated deflection of  10.37 mm.
% of for the case oThese errors increases for  For the case of These errorsFor the case of $T_{\text{base}}=$ 12 N and 16 N and exponent values of $n=$ 2 and 3, these error increases to MAPE $<$ 8 \%, MAE $<$ 1 mm, and $E_{\text{shape}}$ $<$ 1.6 mm increases to $\sim$ 40 \% and $\sim$ 25 \%, respectively. This increase in the error 
% It is important to note that for the case of exponent value $n=$ 1 \yl{and $T_{\text{base}}=$ 12 N and 16 N}, the TD-CM exhibits minimal deformation, with a maximum tip deflection below 10 mm (shown in Fig. \ref{fig:exp2}(a)) \yl{and 12 mm (shown in Fig. \ref{fig:exp2}(b))}. This explains the relatively larger prediction error \yl{(MAPE $\sim$ 40 \% and $\sim$ 25 \%, respectively at $T_{\text{base}}=$ 12 N and 16 N, and MAPE $<=$ 17 \% at $T_{\text{base}}=$ 20 N)}. 
 For the case of larger errors, it is important to acknowledge that  discrepancies between the predicted results and experimental observations can be related to the following  important factors: (1) The resolution and repeatability of the utilized SLS additive manufacturing process and used material to manufacture the mentioned TD-CMs with miniature details; (2) Uncertainties and errors related to the resolution and calibration process of the used motion capture system directly affecting the accuracy of the measured deformations for the TD-CM; and (3) Errors associated with the experimentally-measured values for the friction coefficient and Young's modulus of the fabricated TD-CM's.  Of note, either of these parameters can directly affect on the proposed error metrics. For example, as can be seen in Fig. \ref{fig:exp1}, comparison of the calculated shapes with the experimental outputs  indicates that all of the  additively manufactured TD-CMs consistently experience  a larger deformation compared with their corresponding simulated models. This can be mainly attributed to  the incorrect  parameters (i.e., TD-CM geometrical parameters and Young's modulus) used in the model simulating a stiffer TD-CM  resulting in smaller deformation behaviors. As mentioned, these incorrect values can be related to the fabrication process and the systematic measurement errors related to the procedure and devices utilized to obtain them.

 Also, it is worth emphasizing that, despite the previous literature (e.g., \cite{renda2014dynamic}, \cite{rucker2011statics}, \cite{9707607}), the proposed model simultaneously considers the effect of change in the TD-CMs geometry and friction coefficient on tension loss of a a manipulator with discontinuous geometry. As mentioned, simultaneous change in the geometry and friction coefficient directly affect on the distribution of internal distributed and concentrated loads and subsequently  tension loss and shape of a manipulator that cannot be ignored during the modeling, fabrication, shape sensing, and control of TD-CMs. Therefore, despite the aforementioned potential parametric and measurement errors, the obtained results clearly show the potentials of our proposed approach in modeling such a complicated and nonlinear problem that can be used during the design process of TD-CMs. 

\section{Concluding Remarks}
To conclude, thanks to the introduction of the SEF and  \textit{design space} concepts, this paper presents  a unique geometric tool for the design and comprehensive deformation modeling of a TD-CM with discontinuous geometry experiencing tendon tension loss.  Particularly, by employing the SEF concept, we establish four intuitive functional  mappings relating the manipulator's geometric parameters, kinematics, and deformation behavior enabling efficient design and optimization of TD-CMs with complex FP geometries. Additionally, our model incorporates the effects of geometry and friction coefficient on distributed and concentrated point friction  caused by a partially-constrained tendon within the manipulator's geometry. Of note, this important factor directly affects tension tendon loss and deformation behavior of TD-CMs, which are often overlooked in the existing models (e.g., \cite{camarillo2008mechanics}, \cite{renda20123d}, \cite{boyer2020dynamics}). The performance of the proposed model was evaluated using different simulation and experimental studies. These studies clearly demonstrated  (i) the versatility of the SEF in the design of various TD-CMs simply by changing one exponent value in (\ref{eq:superellipse equation1}); (ii) the effect and importance of FPs and friction coefficients on distribution of internal distributed and concentrated loads, tension loss, and deformation behavior of a partially-constrained tendon in TD-CMs with discrete geometry (shown in Fig. \ref{fig:sim1}, Fig. \ref{fig:sim2}, and Fig.\ref{fig:exp1}) ; (iii) the accuracy of the proposed model in obtaining the shape of three different additively manufactured TD-CMs (see TABLE \ref{table:1}).  

While this study primarily focused on TD-CMs with planar deformation, it is important to note that the proposed framework can be extended to model and evaluate the bending of manipulators in three dimensions, opening avenues for future research.
In future studies, the proposed modeling framework and solution algorithm will also be leveraged to optimize the design of TD-CMs for specific applications such as 
%tendon-driven 
steerable drilling robots for orthopedics applications \cite{alambeigi2017curved,alambeigi2019continuum,sharma2024biomechanics,maroufi2025s3d}.
% By utilizing these tools, researchers can tailor the design parameters to enhance performance, adaptability, and efficiency of TD-CMs in various practical scenarios.
% Therefore, this study provides a solid foundation for future investigations into three-dimensional bending behavior of TD-CMs, and it sets the stage for the application of the proposed modeling framework and solution algorithm in optimizing TD-CM systems to meet specific application requirements.

% \appendix
 \begin{appendices}
\section{}
\label{derivation1}
% For a small material element $ds_c$ along the tendon, the mathematical relationship between the infinitely small change in tendon tension $T$, friction force $F_{\text{friction}}$, and normal force $F_{\text{normal}}$ can be  described as: $dT= dF_{\text{friction}}$ and $dF_{\text{friction}}=\mu dF_{\text{normal}}=\mu Td\theta$.
% Since $f_n=dF_{\text{normal}}/ds_c$
% and $f_f=dF_{\text{friction}}/ds_c$ (forces per unit $s_c$ \cite{rucker2011statics}) and $d\theta/ds_{c}=\kappa_{c}$, we can obtain:
As shown in Fig. \ref{fig:modeling2} and considering a small material element $ds_t$ along the tendon, we can establish the mathematical relationship between the infinitesimal change in tendon tension $dT$, friction force $F_{\text{friction}}$, and normal force $F_{\text{normal}}$ as follows: $dT = dF_{\text{friction}}$ and $dF_{\text{friction}} = \mu dF_{\text{normal}} = \mu T d\theta$.
Here, $f_n = dF_{\text{normal}}/ds_t$ and $f_f = dF_{\text{friction}}/ds_t$ (forces per unit $s_t$ \cite{rucker2011statics}). Additionally, $d\theta/ds_t = \kappa_t$. By utilizing these relationships, we can derive the following expression:
\begin{gather}
\label{eq:normal}
\boldsymbol{f}_n(s)
=\nonumber\underbrace{\boldsymbol{n}_\text{tendon}(s)}_{force\hspace{3pt}direction}\underbrace{\displaystyle\frac{dF_{\text{normal}}}{ds_{t}}}_{force\hspace{3pt}magnitude}\\
=\nonumber\boldsymbol{n}_\text{tendon}(s)\displaystyle\frac{T d\theta}{ds_{t}}=\boldsymbol{n}_\text{tendon}(s)\kappa_t(s) T(s)
\end{gather}
\begin{gather}
\label{eq:friction}
\boldsymbol{f}_f(s)
=\nonumber\underbrace{-\boldsymbol{t}_\text{tendon}(s)}_{force\hspace{3pt}direction}\underbrace{\displaystyle\frac{dF_{\text{friction}}}{ds_{t}}}_{force\hspace{3pt}magnitude}\\
=\nonumber-\boldsymbol{t}_\text{tendon}(s)\displaystyle\frac{\mu T d\theta}{ds_{t}}=-\mu \boldsymbol{t}_\text{tendon}(s) \kappa_t(s)T(s)
\end{gather}

\section{}
\label{derivation2}
As shown in Fig. \ref{fig:modeling2}, at the discontinuous boundary points, the concentrated normal force $\boldsymbol{F}_{N}$ can be calculated by projecting the tendon tension $\boldsymbol{T}$ onto the normal direction $\boldsymbol{n}_\text{tendon}$. Similarly, the concentrated friction force  $\boldsymbol{F}_{F}$ is obtained by projecting the tendon tension $\boldsymbol{T}$ onto the tangent direction $-\boldsymbol{t}_\text{tendon}$. The magnitudes of these projected forces are given by $F_{N}=T\sin{\phi_t}$ and $F_{F}=T \cos{\phi_t}$, respectively. Hence, we can express this relationship as follows:
% $\boldsymbol{F}_{N}=\boldsymbol{n}_\text{tendon}T \sin{\phi_t}$
% $\boldsymbol{F}_{F}=\boldsymbol{t}_\text{tendon}T \cos{\phi_t}$.

\begin{gather}
\begin{cases} \nonumber\boldsymbol{F}_{N}=\underbrace{\boldsymbol{n}_{\text{tendon}}}_{direction} \underbrace{T \sin{\phi_t} }_{magnitude} \\\boldsymbol{F}_{F}=\underbrace{-\boldsymbol{t}_{\text{tendon}}}_{direction} \underbrace{T \cos{\phi_t}}_{magnitude}
\end{cases}
\end{gather}

\end{appendices}

\bibliographystyle{./bib/IEEEtran}

\bibliography{./bib/Soft_robot}

\end{document}